%% file: main.tex
\documentclass[conference]{IEEEtran}
\IEEEoverridecommandlockouts
\usepackage{cite}
\usepackage{amsmath,amssymb,amsfonts}
\usepackage{algorithm}
\usepackage{algorithmic}
\usepackage{pdfpages}
\usepackage{multirow}
\usepackage[export]{adjustbox} 
\usepackage{bm}
\usepackage{amsmath}
\usepackage{array}
\usepackage{booktabs}
\usepackage{footmisc}

\usepackage{amsthm}
\usepackage{amsmath}
\usepackage{amsfonts}
\usepackage{graphicx}
\usepackage{enumitem}

\theoremstyle{Definition}
\newtheorem{mydef}{Definition}
\usepackage{subfigure}
\usepackage{hyperref}
\usepackage{url}
\usepackage{color}
\usepackage{xcolor}
\usepackage{colortbl}
\usepackage{xspace}
\newcommand{\model}{CPDG\xspace}
\input{math_commands.tex}

\usepackage{hyperref}
\usepackage{url}
\usepackage{multirow}
\usepackage{algorithm}
\usepackage{algorithmic}
\usepackage{makecell}

\definecolor{rev}{HTML}{0000CD}

\usepackage{pifont}
\usepackage{enumitem}

\def\BibTeX{{\rm B\kern-.05em{\sc i\kern-.025em b}\kern-.08em
    T\kern-.1667em\lower.7ex\hbox{E}\kern-.125emX}}
\begin{document}

\title{CPDG: A Contrastive Pre-Training Method for Dynamic Graph Neural Networks
}

\author{
    \IEEEauthorblockN{Yuanchen Bei$^{1}$\IEEEauthorrefmark{1}\thanks{* Both authors contributed equally to this paper.}, Hao Xu$^{2}$\IEEEauthorrefmark{1}, Sheng Zhou$^{1,3}$\IEEEauthorrefmark{2}\thanks{† Corresponding author.}, Huixuan Chi$^4$,}
    \IEEEauthorblockN{Haishuai Wang$^1$, Mengdi Zhang$^2$, Zhao Li$^1$, Jiajun Bu$^1$\IEEEauthorrefmark{2}}
    \IEEEauthorblockA{$^1$ Zhejiang Provincial Key Laboratory of Service Robot, College of Computer Science and Technology, Zhejiang University}
    \IEEEauthorblockA{$^2$ Meituan $\quad$ $^3$ School of Software Technology, Zhejiang University}
    \IEEEauthorblockA{$^4$ Institute of Computing Technology, Chinese Academy of Science}
    \IEEEauthorblockA{yuanchenbei@zju.edu.cn, kingsleyhsu1@gmail.com, 
    zhousheng\_zju@zju.edu.cn,
    chihuixuan21s@ict.ac.cn,} 
    \IEEEauthorblockA{haishuai.wang@gmail.com, 
    mdzhangmd@gmail.com,
    zhao\_li@zju.edu.cn,
    bjj@zju.edu.cn}
}

\maketitle

\begin{abstract}
\input{abstract.tex}
\end{abstract}

\begin{IEEEkeywords}
dynamic graph neural networks, pre-training, contrastive learning
\end{IEEEkeywords}

\section{Introduction}
\input{introduction.tex}

\section{Related Works} \label{sec:related}

\input{related.tex}

\begin{figure*}[t]
    \centering
    \includegraphics[width=\textwidth, trim=0cm 0cm 0cm 0cm,clip]{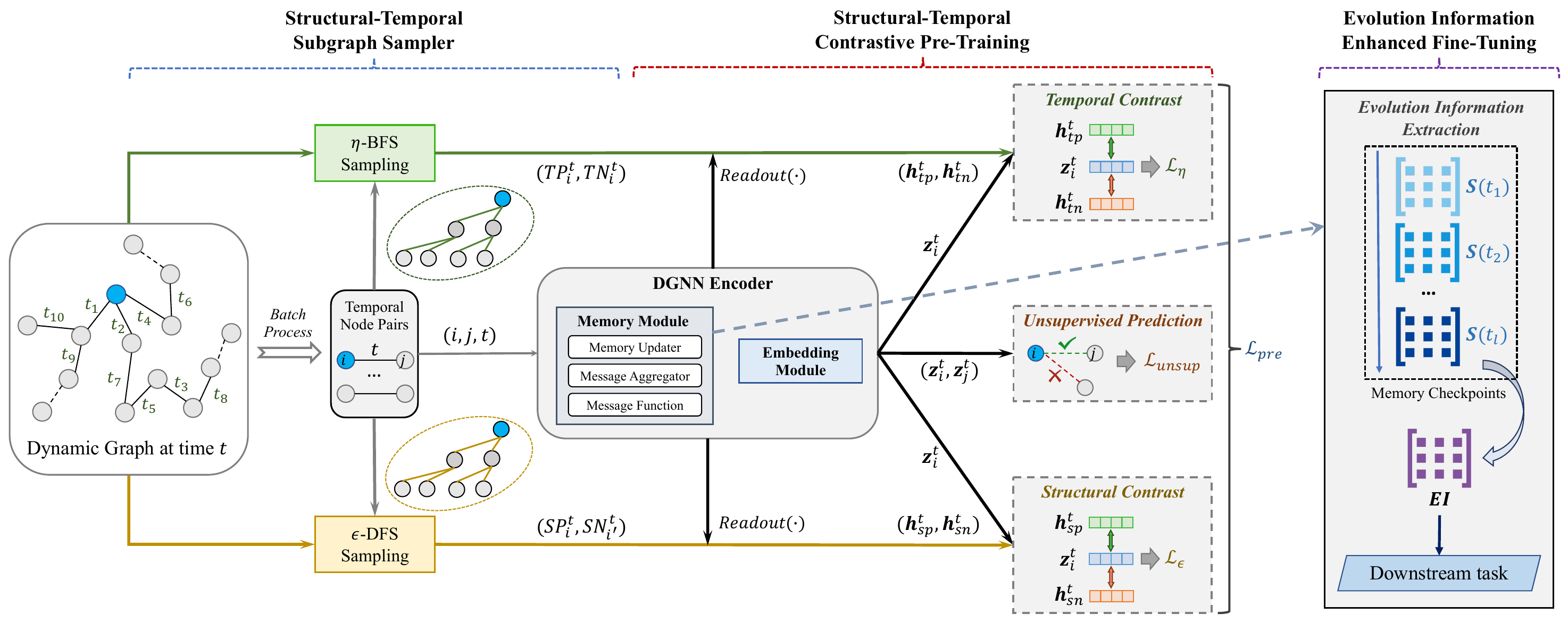}
    \caption{The overall framework of the proposed \model model. \model contains three major parts:
    (i) the structural-temporal subgraph sampler which obtains the positive-negative subgraph pairs with the temporal preference by $\eta$-BFS sampling and with the structural preference by $\epsilon$-DFS sampling;
    (ii) structural-temporal contrastive pre-training which mines the informative and transferable long-short term temporal patterns from temporal contrast and structural patterns from structural contrast; 
    (iii) evolution information enhanced fine-tuning which is an optional module that can assist fine-tuning in downstream tasks with the pre-trained evolution information.}
    \label{fig:framework}
    \vspace{-0.8em}
\end{figure*}

\section{Preliminaries} \label{sec:pre}
\input{preliminaries.tex}

\section{Methodology} \label{sec:method}

\input{method.tex}

\section{Experiments} \label{sec:exper}
\input{experiment_new.tex}

\vspace{-0.69em}

\section{Conclusion} \label{sec:con}
\input{conclusion.tex}

\section*{Acknowledgments}
This work is supported in part by the National Natural Science Foundation of China (Grant No. 62106221, 62372408),
Zhejiang Provincial Natural Science Foundation of China
(Grant No. LTGG23F030005), and Ningbo Natural Science
Foundation (Grant No. 2022J183).

\bibliographystyle{IEEEtran}
\bibliography{IEEEabrv,ref_new}


\end{document}

%% file: math_commands.tex
\usepackage{amsmath,amsfonts,bm}

\def\eqref#1{equation~\ref{#1}}

\def\1{\bm{1}}

\DeclareMathAlphabet{\mathsfit}{\encodingdefault}{\sfdefault}{m}{sl}
\SetMathAlphabet{\mathsfit}{bold}{\encodingdefault}{\sfdefault}{bx}{n}

\def\gE{{\mathcal{E}}}

\def\gG{{\mathcal{G}}}

\def\gL{{\mathcal{L}}}

\def\gN{{\mathcal{N}}}
\def\gO{{\mathcal{O}}}

\def\gV{{\mathcal{V}}}



%% file: abstract.tex
Dynamic graph data mining has gained popularity in recent years due to the rich information contained in dynamic graphs and their widespread use in the real world. Despite the advances in dynamic graph neural networks (DGNNs), the rich information and diverse downstream tasks have posed significant difficulties for the practical application of DGNNs in industrial scenarios.
To this end, in this paper, we propose to address them by pre-training and present the \underline{C}ontrastive \underline{P}re-Training Method for \underline{D}ynamic \underline{G}raph Neural Networks (\model). 
\model tackles the challenges of pre-training for DGNNs, including \textit{generalization} capability and \textit{long-short term modeling} capability, through a flexible structural-temporal subgraph sampler along with structural-temporal contrastive pre-training schemes. Extensive experiments conducted on both large-scale research and industrial dynamic graph datasets show that \model outperforms existing methods in dynamic graph pre-training for various downstream tasks under three transfer settings.

%% file: introduction.tex
Graph is ubiquitous in real-world scenarios and mining graph data has attracted increasing attention from both academic and industry communities~\cite{ying2018graph,wu2020comprehensive,bei2023reinforcement}, such as epidemiology~\cite{pastor2001epidemic,may2001infection}, bioinformatic~\cite{li2019deep,jumper2021highly}, and recommender system~\cite{wu2020graph,he2020lightgcn,bei2023non}. 
Previous graph mining works have mainly focused on \textit{static} graphs, neglecting the \textit{dynamic} nature of real-world graph data, especially in industrial scenarios~\cite{skarding2021foundations}.
For example, in Meituan\footnote{\url{https://meituan.com}}, users engage in real-time interactions with various items and shops, forming a rapidly changing large-scale dynamic graph.
Such dynamic and evolving patterns are essential for accurate graph modeling~\cite{xu2020inductive}, which has gained growing interest in practice.

Recently, the dynamic graph neural networks~(DGNNs) have achieved success for dynamic graph mining~\cite{nguyen2018continuous,kumar2018learning,trivedi2019dyrep,rossi2020temporal,xu2020inductive,you2022roland}.
A typical DGNNs workflow trains the model on dynamic graph data and then directly applies it for inference~\cite{skarding2021foundations}.
However, the learned patterns from training may not always be applicable to new inference graphs, especially in dynamic and large-scale industrial systems.
Furthermore, frequent retraining is especially impractical in large-scale industrial systems, where billions of interactions may occur within a short time interval.
On the one hand, DGNNs require simultaneous consideration of temporal and structural information, resulting in higher computational complexity compared to traditional graph neural networks (GNNs) for static graphs.
On the other hand, the unique requirements of different tasks in dynamic graphs, such as node-level or graph-level and temporal or structural, further limit the possibility of retraining for each new task.
To sum up, \textit{rich information and various downstream tasks have posed significant difficulties for the application of DGNNs}.

In recent years, the \textit{pre-training and fine-tuning} strategy has been widely studied in natural language processing~\cite{devlin2019bert}, computer vision~\cite{khan2022transformers} and gradually expanded to the graph domain~\cite{hu2019strategies,qiu2020gcc,hu2020gpt,lu2021learning,sun2022gppt}.
The goal of graph pre-training is to learn the transferable knowledge representation on large-scale graph data, which can be applied to various downstream applications and tasks.
Coincidentally, such a strategy has the potential to address the aforementioned challenges in applying DGNNs to real-world systems through pre-training on historical dynamic graph data, followed by fine-tuning for specific downstream tasks, rather than retraining from scratch.

Although feasible, the dynamic nature has brought extra challenges to pre-training DGNNs compared with static graph neural networks:
First, \textbf{Generalization Capability}. Generalization is fundamental but essential for pre-training strategies~\cite{liu2023graphprompt}, particularly in cases where dynamic information brings various downstream tasks that involve both temporal and structural aspects.
However, current DGNNs are mostly trained for specific tasks like dynamic link prediction and lack the ability to generalize to various downstream tasks~\cite{xu2020inductive,rossi2020temporal}.
Second, \textbf{Long-Short Term Modeling Capability}.
In dynamic graphs, both long-term stable patterns and short-term fluctuating patterns are important natures and urgent for downstream tasks, especially for rapidly changing industrial dynamic graphs.
Existing DGNNs have focused on modeling consistent changes over time to capture the long-term stable temporal patterns by utilizing memory buffers or temporal smoothing operators~\cite{rossi2020temporal,tian2021self}.
However, when pre-training on graphs with large time intervals, the short-term fluctuating patterns can easily be overshadowed by long-term stable patterns.
How to simultaneously capture both long-short term patterns during DGNN pre-training is still under-explored.

\begin{figure}[tbp]
    \centering
    \includegraphics[width=0.95\linewidth, trim=0cm 0cm 0cm 0cm,clip]{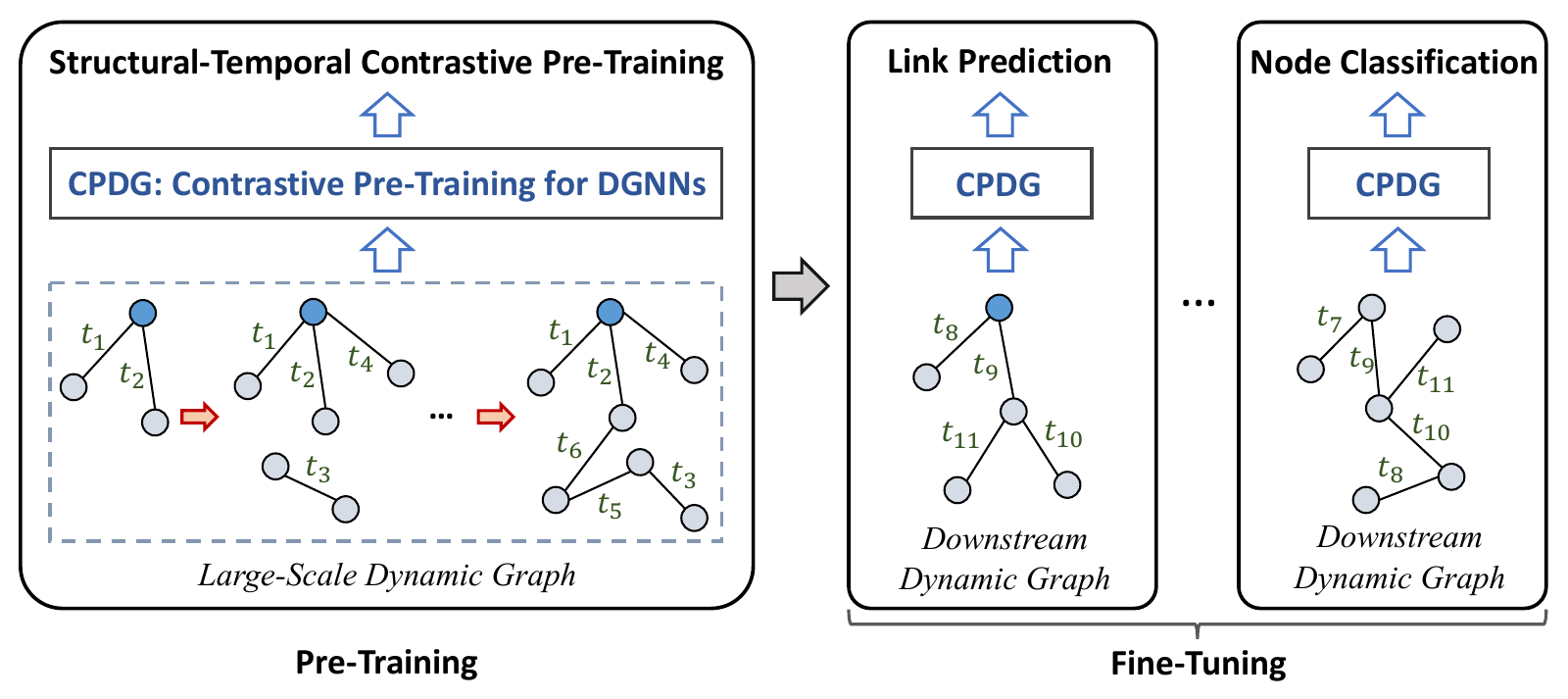}
    \caption{Overall pre-training and fine-tuning workflow of the proposed \model.}
    \label{fig:workflow}
    \vspace{-0.6em}
\end{figure}

To tackle the above challenges, in this paper, we propose \textbf{\model}, a novel \underline{C}ontrastive \underline{P}re-Training Method for \underline{D}ynamic \underline{G}raph Neural Networks.
Figure \ref{fig:workflow} illustrates the overall workflow of our proposed \model method.
More specifically, \model first proposes a structural-temporal sampler that extracts informative subgraph with flexible temporal-aware sampling probability.
Additionally, both the temporal and structural contrastive pre-training schemes are carefully designed to learn transferable long-short term evolution patterns and discriminative structural patterns in dynamic graphs.
To further benefit the downstream tasks, \model explicitly extracts the evolved patterns during pre-training and provides for downstream fine-tuning.
We conduct extensive experiments on both large-scale real-world dynamic graph datasets and the industrial dataset in Meituan. The results demonstrate that \model outperforms state-of-the-art graph pre-training methods on all datasets under various transfer settings and downstream tasks. 
Further experimental studies of \model prove the effectiveness of the designed method from multiple dimensions. The main contributions of this paper are summarized as follows.
\begin{itemize}[leftmargin=*]
    \item We highlight the difficulties of applying DGNNs in industrial scenarios,  that the model complexity and various tasks make it impractical to retrain DGNNs.
    We propose to address them with pre-training for DGNNs which has received limited attention in the literature.
    \item We propose a novel \model method to tackle the challenges of pre-training and efficiently learn the transferable knowledge on large-scale dynamic graphs by a novel flexible structural-temporal subgraph sampler along with two views of subgraph contrastive pre-training schemes.
    \item We conduct extensive experiments on both large-scale public dynamic graph datasets and the industrial dataset in Meituan with different transfer settings and downstream tasks. Experimental results demonstrate the effectiveness and the generalization ability of \model.
\end{itemize}




In the following sections, we will first review the previous work related to our method in Section \ref{sec:related}. Second, we will give some key preliminaries of our work in Section \ref{sec:pre}.
Then, the detailed description of the proposed \model will be introduced in Section \ref{sec:method}.
To further verify the effectiveness of \model, we conduct various experiments in Section \ref{sec:exper}.
Finally, the conclusion of this paper is posed in Section \ref{sec:con}.

%% file: related.tex
\begin{table}[tbp]
  \centering
  \caption{Characteristics Comparison between representative pre-training methods and our proposed \model.}
  \resizebox{\linewidth}{!}{
    \begin{tabular}{l|cccc}
    \toprule
    Key Properties & \makecell[c]{Dynamic \\Data}	 & \makecell[c]{Long-term \\Stable Pattern} & \makecell[c]{Short-term \\Fluctuating Pattern} & \makecell[c]{Pre-training \\Strategy} \\
    \midrule
    DGI~\cite{velickovic2019deep}   &  \textcolor{red}{×}    &   \textcolor{red}{×}     &   \textcolor{red}{×}    & Contrastive \\
    GPT-GNN~\cite{hu2020gpt} &  \textcolor{red}{×}  &   \textcolor{red}{×}     &    \textcolor{red}{×}   & Generative \\
    PT-DGNN~\cite{chen2022pre} &   \textcolor[HTML]{006400}{\checkmark}     &  \textcolor{red}{×}      &   \textcolor{red}{×}    & Generative \\
    DDGCL~\cite{tian2021self} &   \textcolor[HTML]{006400}{\checkmark}    &   \textcolor[HTML]{006400}{\checkmark}    &   \textcolor{red}{×}    & Contrastive \\
    SelfRGNN~\cite{sun2022self} &   \textcolor[HTML]{006400}{\checkmark}    &    \textcolor[HTML]{006400}{\checkmark}   &   \textcolor{red}{×}    & Contrastive \\
    \midrule
    \textbf{CPDG} (ours) &   \textcolor[HTML]{006400}{\checkmark}    &  \textcolor[HTML]{006400}{\checkmark}     &  \textcolor[HTML]{006400}{\checkmark}      & Contrastive \\
    \bottomrule
    \end{tabular}%
    }
  \label{tab:model_comp}%
\end{table}%

\subsection{Dynamic Graph Neural Networks}
Graph Neural Networks (GNNs) have shined a light on graph-structured data mining in recent years to guide the representation learning~\cite{zhang2020deep,kipf2016semi,hamilton2017inductive,velivckovic2018graph,wu2020comprehensive,zhang2023alleviating}. 
However, since these networks are mainly designed for \textit{static graphs}, they lack the ability to simultaneously capture the temporal and structural patterns within dynamic graphs, which are more common in the real world. Therefore, Dynamic Graph Neural Networks~(DGNNs) have gained increasing attention recently, and many methods have been proposed and achieved great success in dynamic graph representation learning~\cite{nguyen2018continuous,kumar2018learning,trivedi2019dyrep,rossi2020temporal,wang2020inductive,xu2020inductive,you2022roland}.

Among them, DyRep~\cite{trivedi2019dyrep} posits dynamic graph representation learning as a latent mediation process and utilizes the temporal point process for dynamic graph modeling. JODIE~\cite{kumar2018learning} designs a coupled recurrent model to learn dynamic node embeddings with the update, projection, and prediction components. TGAT~\cite{xu2020inductive} further designs a self-attention temporal-topological neighborhood aggregator for dynamic graph modeling. 
Then, TGN~\cite{rossi2020temporal} designs a generic and state-of-the-art framework for dynamic graph learning with the memory module and unifies most of the above methods into this framework.


\subsection{Pre-Training for GNNs}
With the continuous success of GNNs in graph mining and the trend of increasing graph data scale in recent years, a growing number of works have begun to note the important role of graph pre-training and explore various solutions for pre-training GNNs on unannotated graph data~\cite{hu2019strategies,qiu2020gcc,hu2020gpt,lu2021learning,jiang2021pre,sun2022gppt}.

Most current works focus on static graphs. Among them, a direct way is to design some unsupervised tasks to let GNNs (e.g. GraphSAGE~\cite{hamilton2017inductive} and GAT~\cite{velivckovic2018graph}) pre-train on unlabeled data, such as link prediction~\cite{lichtenwalter2010new}.
Another category of methods is graph contrastive learning-based pre-training: DGI~\cite{velickovic2019deep} maximizes the mutual information between node representations and graph summaries. Recently, GCC~\cite{qiu2020gcc} proposes a contrastive learning scheme on subgraph perspectives with an instance discrimination task to learn transferable universal structural patterns.
Besides, GPT-GNN~\cite{hu2020gpt} designs a generative graph pre-training paradigm with masked node attribute generation and edge generation tasks inspired by the pre-train language model.

In recent years, few works have begun to pay attention to pre-training DGNNs on dynamic graphs. PT-DGNN~\cite{chen2022pre} improves GPT-GNN with temporal-aware masking. DDGCL~\cite{tian2021self} maximizes the time-dependent agreement between a node identity's two temporal views. Recently, SelfRGNN~\cite{sun2022self} proposes a Riemannian reweighting self-contrastive approach for self-supervised learning on dynamic graphs.
We illustrate and compare the key characteristics between our designed \model and these state-of-the-art pre-training methods in Table \ref{tab:model_comp}.

%% file: preliminaries.tex
In this section, we present the key definitions related to pre-training for DGNNs. The main symbols are listed in Table \ref{tab:symbol}.

\subsection{Dynamic Graph}
There exist two main types of dynamic graphs, \textit{discrete-time dynamic graphs} (DTDG) and \textit{continuous-time dynamic graphs} (CTDG)~\cite{skarding2021foundations}.
DTDG is a sequence of static graph snapshots taken at intervals in time. CTDG is more general and can be represented as temporal lists of events, which reflects more detailed temporal signals and fine-grained rather than the coarse-grained DTDG.
In this paper, we focus on the CTDG that is widely used in industrial systems and more demand for pre-training, which can be formulated as:

\begin{mydef}
\textbf{Dynamic Graph}~\cite{rossi2020temporal} is defined as $\gG=(\gV^{T}, \gE^{T})$, where $\gV^{T}$ is a temporal set of vertices, $\gE^{T}$ is the temporal set of edges, and $T$ is the time set.
$N=|\gV^T|$ denotes the number of vertices in $\gG$.
Each edge $e_{i, j}^{t} \in \mathcal{E}^{T}$ is denoted as a triple, $e_{i, j}^{t} = (i, j, t)$, where node $i, j \in \mathcal{V}^T$ and time $t \in T$.
Each $(i, j, t)$ means node $i$ has an interaction with node $j$ at time $t$.
We denote the temporal graph $\gG$ at time $t$ as the graph $\gG^t = (\gV^t, \gE^t)$, $\gV^t$ and $\gE^t$ as the set of vertices and edges observed before time $t$.
We denote $\gN_i^t = \{j\ |\ e_{i, j}^{t-} \in \gE^t, t^- \le t\} $ as the neighborhood set of node $i$ in time interval $[0, t]$, and $\gN_i^{k,t}$ as the set of k-hop neighborhoods of node $i$. 
\end{mydef}

\subsection{Dynamic Graph Neural Networks}
Given a dynamic graph $\gG=(\gV^{T}, \gE^{T})$, the existing Dynamic Graph Neural Network (DGNN) encoder learns the temporal embeddings for all nodes at time $t$, denoted as $\mathbf{Z}^t = (\mathbf{z}_1^t, ..., \mathbf{z}_{i}^t, ...)$.
Besides, a memory $\mathcal{M}$ stores the memory state for each node at time $t$, denoted as $\mathbf{S}^t = (\mathbf{s}_1^t, ..., \mathbf{s}_{i}^t, ...)$, in which each state memorizes the temporal evolution of the node at the time interval $[0, t]$ in compressed format. 
The paradigm of DGNN encoder can be formulated as a function of node $i$ and its k-hop temporal neighbors $\mathcal{N}_i^{k, t}$ at time $t$: 
\begin{equation}
    \mathbf{z}_{i}^t = \mathrm{Emb}(i, t) = \sum_{u\in\gN_{i}^{k, t}}f\left(\mathbf{s}_{i}^t, \mathbf{s}_{u}^{t-}\right),
\end{equation}
where $f(\cdot)$ denotes a learnable function \cite{rossi2020temporal}, such as dynamic graph attention. $\mathbf{s}_{i}^t$ denotes the state of node $i$ that is stored in memory at time $t$, and $\mathbf{s}_{u}^{t{-}}$ is the latest state of node $u$ before time $t$, which is initialized as a zero vector for new encountered nodes and updated with batch processing by three following steps: \textit{Message Function}, \textit{Message Aggregator} and \textit{Memory Updater}.

\textit{(i) Message Function}. Given an interaction $(i,j,t)$ involving node $i$, a message is computed to update state $\mathbf{s}_{i}^t$ of node $i$ in the memory at time $t$, which can be expressed as:
\begin{equation}
    \mathbf{m}_{i}^t = \mathrm{Msg}\left(\mathbf{s}_{i}^{t{-}}, \mathbf{s}_{j}^{t{-}}, \phi(\Delta t)\right),
\end{equation}
where $\mathrm{Msg}(\cdot)$ is the message function, such as identity and MLP~\cite{rossi2020temporal}. 
$\Delta t$ denotes the last updated time interval between $\mathbf{s}_{i}^{t{-}}$ and $\mathbf{s}_{j}^{t{-}}$, $\phi(\cdot)$ represents a generic time encoding~\cite{xu2020inductive}. 

\textit{(ii) Message Aggregator}. 
As each node $i$ may have multiple interaction events in time interval $[t_1, t]$ and each event generates a message, a message aggregator is further used to aggregate all the messages $\mathbf{m}_{i}^{t_{1}}, ..., \mathbf{m}_{i}^{t_{b}}$ for $t_1, ..., t_b \le t$:
\begin{equation}
    \mathbf{\overline{m}}_{i}^t = \mathrm{Agg}\left(\mathbf{m}_{i}^{t_{1}}, ..., \mathbf{m}_{i}^{t_{b}}\right),
\end{equation}
where $\mathrm{Agg}(\cdot)$ is an aggregation function, such as mean and last time aggregation \cite{rossi2020temporal}.

\textit{(iii) Memory Updater}. 
The state $\mathbf{s}_{i}^t$ of node $i$ at time $t$ is updated upon $\mathbf{\overline{m}}_{i}^t$ and its previous state $\mathbf{s}_{i}^{t{-}}$, which can be formulated as:
\begin{equation}
    \mathbf{s}_{i}^t = \mathrm{Mem}\left(\mathbf{s}_{i}^{t{-}}, \mathbf{\overline{m}}_{i}^t\right),
\end{equation}
where $\mathrm{Mem}(\cdot)$ represents a time series function, such as RNN \cite{medsker2001recurrent}, LSTM \cite{hochreiter1997long} and GRU \cite{chung2014empirical}.

Note that most popular DGNN encoders have followed the above paradigm, such as JODIE~\cite{kumar2018learning}, DyRep~\cite{trivedi2019dyrep} and TGN~\cite{rossi2020temporal}. They differ in the implementation of $f(\cdot)$, $\mathrm{Msg}(\cdot)$, $\mathrm{Agg}(\cdot)$ and $\mathrm{Mem}(\cdot)$, as compared in Table \ref{tab:backbone_comp} following the organization in previous works~\cite{rossi2020temporal}.

\begin{table}[tbp]
  \centering
  \caption{Main symbols and definitions in the paper.}
  \resizebox{\linewidth}{!}{
    \begin{tabular}{c|c}
    \toprule
    \multicolumn{1}{c|}{Notations} & Description \\
    \midrule
    \midrule
     $\mathcal{G}=(\mathcal{V}^{T}, \mathcal{E}^{T})$     & A dynamic graph. \\
     \hline
     $\mathcal{V}^{T}$ & The temporal set of vertices. \\
     \hline
     $\mathcal{E}^{T}$     & The temporal set of edges. \\
     \hline
     $T$    & The time set. \\
     \hline
     $N=|\gV^T|$ & The number of vertices. \\
     \hline
     $\mathcal{G}^{t}=(\mathcal{V}^{t}, \mathcal{E}^{t})$     & The dynamic graph at time $t$. \\
     \hline
     $\mathcal{V}^{t}$     & The set of vertices observed before time $t$. \\
     \hline
    $\mathcal{E}^{t}$     & The set of edges observed before time $t$. \\
    \hline
      $\mathcal{N}^{t}_{i}$    & The neighborhood set of node $i$ before time $t$. \\
      \hline
      $\mathcal{N}^{k,t}_{i}$ & The k-hop neighborhoods of node $i$ before time $t$. \\
      \hline
      $\mathbf{Z}^{t}$    & The temporal embeddings at time $t$. \\
      \hline
      $\mathbf{S}^{t}$    & The memory slots at time $t$.\\
      \hline
      $\mathbf{m}^{t}_{i}$ & A message to node $i$ at time $t$. \\
      \hline
      $T^{t}_{i}$    & The set of event time that contains node $i$ by time $t$. \\
      \midrule
    \bottomrule
    \end{tabular}%
    }
    \vspace{-0.3em}
  \label{tab:symbol}%
\end{table}%

\begin{table}[tbp]
  \centering
  \caption{Implement for $f(\cdot)$, $\mathrm{Msg}(\cdot)$, $\mathrm{Agg}(\cdot)$ and $\mathrm{Mem}(\cdot)$ of different DGNN encoders.}
  \resizebox{0.91\linewidth}{!}{
    \begin{tabular}{l|cccc}
    \toprule
    Model & $f(\cdot)$ & $\mathrm{Msg}(\cdot)$ & $\mathrm{Agg}(\cdot)$ & $\mathrm{Mem}(\cdot)$  \\
    \midrule
    JODIE~\cite{kumar2018learning} & Time & Identity & -- & RNN \\
    DyRep~\cite{trivedi2019dyrep} & Identity & Attention & -- & RNN \\
    TGN~\cite{rossi2020temporal} & Attention & Identity & Last-time & GRU \\
    \bottomrule
    \end{tabular}%
    }
  \label{tab:backbone_comp}%
\end{table}%

\begin{mydef}
\textbf{Pre-Training for Dynamic Graph Neural Networks} aims at pre-training a generalizable DGNN encoder $f_{\theta}$ on unlabeled large-scale dynamic graph $\gG=(\gV^{T}, \gE^{T})$ in a self-supervised way so that $f_{\theta}$ can benefit  initialization of models in downstream tasks.
It is worth noting that in the dynamic graph downstream tasks, the newly encountered events may contain nodes that occur in the pre-training stage, thus it is also beneficial to extract the temporal pattern of these nodes and propagate them in the fine-tuning stage.
\end{mydef}

%% file: method.tex
In this section, we introduce our proposed \textbf{\model}, which consists of the following parts:
Firstly, a flexible \textit{\textbf{structural-temporal subgraph sampler}} is applied with various sampling strategies to obtain subgraphs with different preferences. More specifically, we provide a $\eta$-BFS sampling strategy and a $\epsilon$-DFS sampling strategy for extracting the subgraph with temporal preference and structural preference.
Secondly, with the proposed sampler, a \textit{\textbf{structural-temporal contrastive pre-training}} module is further proposed with both temporal contrast and structural contrast to learn transferable long-short term evolution patterns and discriminative structural patterns simultaneously.
Finally, an optional \textit{\textbf{evolution information enhanced fine-tuning}} module is further proposed to extract informative long-short term evolution patterns during pre-training and provide for downstream tasks.
Figure \ref{fig:framework} illustrates the overall network architecture of \model. 

\subsection{Structural-Temporal Subgraph Sampler} 
To obtain informative and transferable information for model pre-training, we design two sampling strategies for different focuses rather than uniformly random sampling schemes in most existing methods~\cite{rossi2020temporal,xu2020inductive}.
The structural-temporal subgraph sampler is designed to flexibly access various structural/temporal-aware subgraph sampling functions on the dynamic graph data for extracting subgraphs with structural-temporal preferences and providing contrastive pre-training with informative samples.

\subsubsection{\textbf{$\eta$-BFS Sampling Strategy}}\label{e-bfs}

Existing works on DGNNs have successfully modeled the long-term stable patterns within dynamic graphs by utilizing memory buffers or temporal smoothing restrictions~\cite{rossi2020temporal,tian2021self,sun2022self}.
However, in addition to the long-term stable patterns that can be modeled by current DGNN models, the short-term fluctuating patterns are also of great significance for real-life industrial graphs with rapid changes.
In real-world recommender systems, both the long-term and short-term interests of users extracted from the user-item interaction graph are important for making optimal recommendation decisions~\cite{yu2019adaptive,chi2023modeling}.
For instance, in the Meituan platform, besides the enduring stable trend of users' personal taste preferences, popular factors contribute to the rapid daily fluctuations in user interests~\cite{sun2022graph}.
In order to further capture the short-term fluctuating temporal evolution patterns in dynamic graphs during pre-training, we design a $\eta$-BFS sampling strategy to extract temporal subgraphs by accessing various temporal-aware sampling functions.

\begin{figure}[tbp]
    \centering
    \includegraphics[width=\linewidth, trim=0cm 0cm 0cm 0cm,clip]{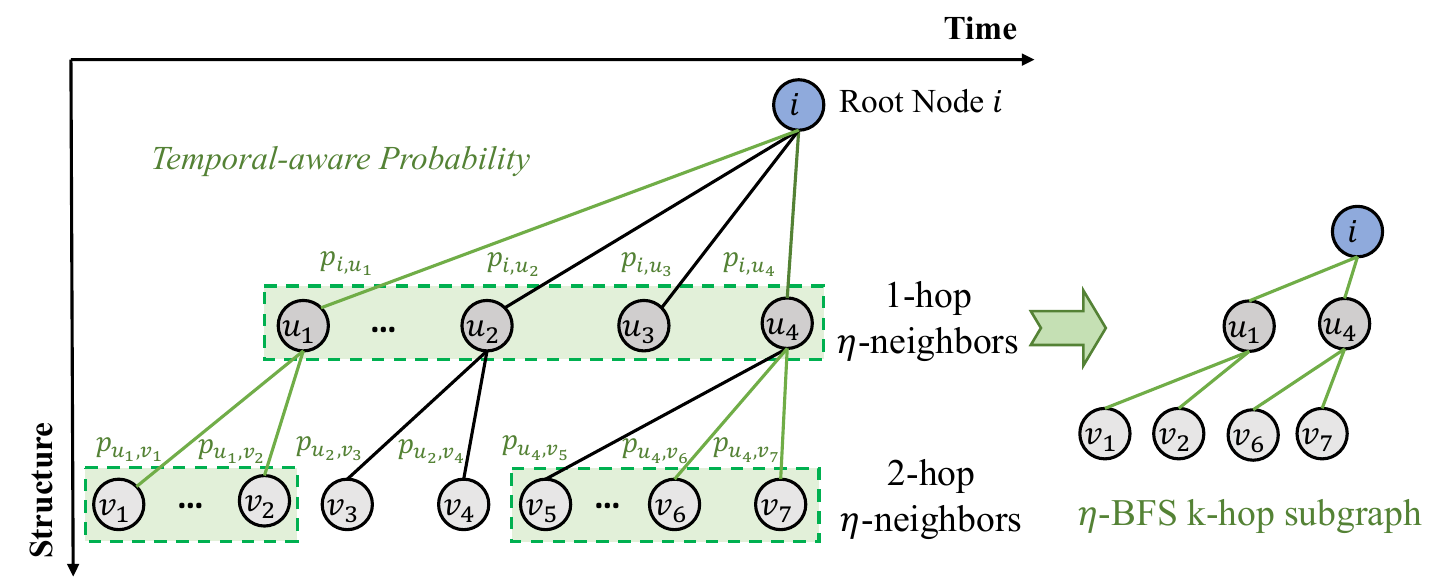}
    \caption{A toy example of $\eta$-BFS sampling with $\eta=2$ and $k=2$. Node $u_1$ and $u_4$ are first selected as 1-hop $\eta$-neighbors, $\{v_1,v_2,v_6,v_7\}$ are then sampled from the neighborhoods of $u_1$ and $u_4$ as 2-hop $\eta$-neighbors, the output of $\eta$-BFS sampling is a subgraph consisting of $\{i,u_1,u_4,v_1,v_2,v_6,v_7\}$.
    }
    \vspace{-1.em}
    \label{fig:bfs}
\end{figure}

We first define the set of event time that contains node $i$ by time $t$ as $T_{i}^{t} = \{t_u | (i, u, t_u) \in\gE^{t}, t_u < t\}$, given a root node $i$ at time $t$, the $\eta$-BFS sampling first observes all the 1-hop neighbors $u\in \gN_i^t$ of node $i$. A sampling probability $p_{u}$ is then assigned to each neighbor $u$ by a temporal-aware function $f_{t\rightarrow p}(\cdot)$ under $T_{i}^{t}$.
The 1-hop $\eta$-neighbors are randomly sampled from $\gN_i^t$ depending on $p_{u}$.
By conducting the above sampling process on each sampled 1-hop $\eta$-neighbors, we get the 2-hop $\eta$-neighbors.
After recursively conducting the above sampling for $k$ times, the $\eta$-BFS sampling generates the $\eta$-BFS $k$-hop subgraph.
Figure \ref{fig:bfs} illustrates a toy example of $\eta$-BFS sampling with $\eta=2$ and $k=2$.
The $\eta$-BFS sampling strategy will be utilized with two designed temporal-aware probability functions to generate sample pairs for the following temporal contrastive learning in Subsection \ref{pretrain-downstream}.

\subsubsection{\textbf{$\epsilon$-DFS Sampling Strategy}}
In addition to the temporal evolution pattern, the unique and discriminative structural pattern of each node is also indispensable for the pre-training of DGNNs~\cite{qiu2020gcc}.
A naive way to capture the unique structural pattern is conducting a random walk starting from node $i$ to generate the subgraph~\cite{perozzi2014deepwalk}, which is equivalent to the depth-first search (DFS) rooted by node $i$.
However, such vanilla DFS will miss the temporal order that evolved over time in dynamic graphs.
We further extend the vanilla DFS with temporal-aware selection and propose a structural $\epsilon$-DFS sampler for maintaining both structural and temporal characteristics.

\begin{figure}[tbp]
    \centering
    \includegraphics[width=\linewidth, trim=0cm 0cm 0cm 0cm,clip]{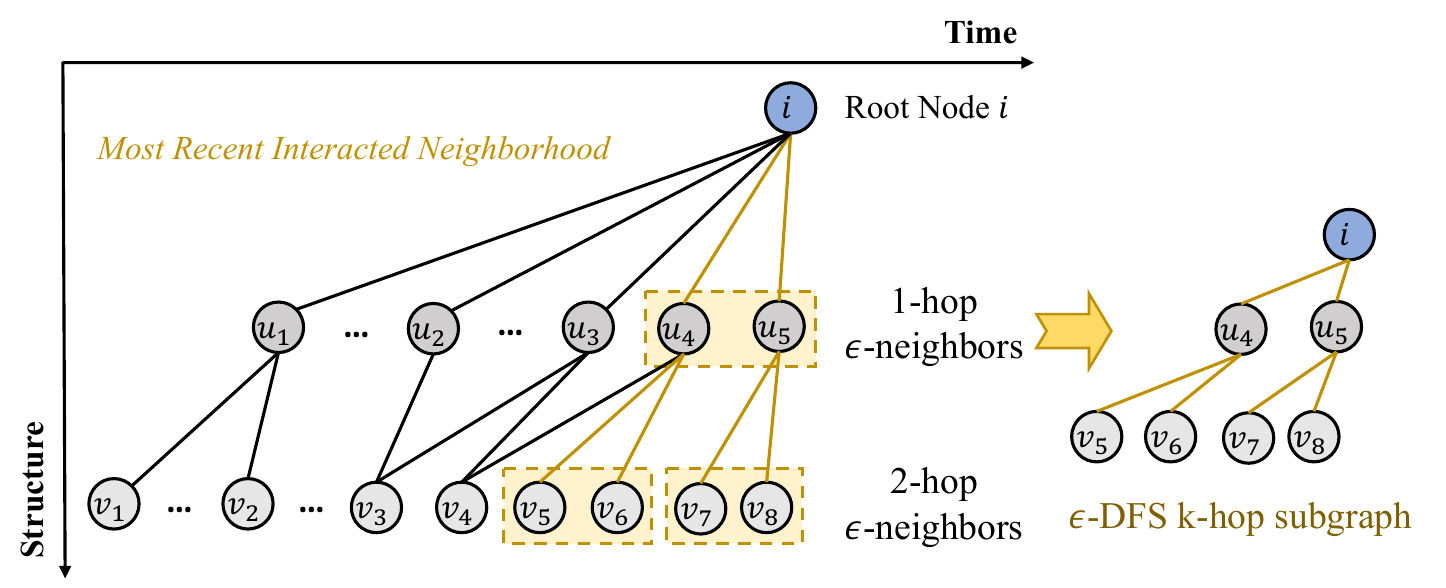}
    \caption{A toy example of $\epsilon$-DFS sampling with $\epsilon=2$ and $k=2$. Node $u_4$ and $u_5$ are the latest interacted two nodes and thus selected as 1-hop $\epsilon$-neighbors, $\{v_5,v_6,v_7,v_8\}$ are the latest interacted nodes of $\{u_4,u_5\}$ and selected as 2-hop $\epsilon$-neighbors, the output of $\epsilon$-DFS sampling is a subgraph consisting of $\{i,u_4,u_5,v_5,v_6,v_7,v_8\}$.}
    \label{fig:dfs}
    \vspace{-1.em}
\end{figure}

Following the definition of $T_{i}^{t} = \{t_u | (i, u, t_u) \in\gE^{t}, t_u < t\}$ in Subsection \ref{e-bfs}, we chronologically sort all the 1-hop neighbors from $\gN_{i}^t$ as $\mathcal{NS}_{i}^t$, and select the most recent interacted $\epsilon$ neighbors:
\begin{align}
    \mathcal{NS}_{i}^t = [a, b, ..., \underbrace{u, ..., v, ...}_{\epsilon-\mathrm{neighbors}}],
\end{align}
where $t_{a} \le t_{b} \le ... \le t_{u} \le t_{v} \le ... < t$. 
It is worth noting that such expansion only considers the most recent temporal information, which is consistent with the $\eta$-BFS but in a discrete formulation.
Then, for each selected 1-hop neighbor, we add it to subgraph $\gG_{i}^{t}$ and repeat the above sampling process for $k$ times. 
Figure \ref{fig:dfs} illustrates a toy example of $\epsilon$-DFS sampler with $\epsilon=2$ and $k=2$.
The $\epsilon$-DFS sampling strategy will be utilized to generate sample pairs for the following structural contrastive learning in Subsection \ref{pretrain-downstream}.

Since both $\eta$-BFS and $\epsilon$-DFS samplers are independent of the learnable parameters or node embeddings during the pre-training stage, the sampling 
operators can be preprocessed before pre-training for efficiency.

\subsection{Structural-temporal Contrastive Pre-training}\label{pretrain-downstream}
With the help of the proposed structural-temporal subgraph sampler with the designed $\eta$-BFS sampling strategy and $\epsilon$-DFS sampling strategy, 
the structure-temporal contrastive pre-training is designed to capture the informative and transferable long-short term evolution patterns and structural patterns from these samples.

\subsubsection{\textbf{Temporal Contrast (TC)}}
The DGNN backbone encoders with the memory module have the ability to learn long-term stable patterns~\cite{kumar2018learning,trivedi2019dyrep,rossi2020temporal}.
Nevertheless, the short-term fluctuating evolution patterns over time are also vital for DGNNs' real-world applications, such as user modeling in above mentioned industrial scenarios.
Thus, short-term fluctuations should also be highlighted during the modeling at each timestamp.
Therefore, the temporal contrast is proposed by contrasting the recent subgraph (positive sample) with the agelong subgraph (negative sample) to capture the short-term fluctuating patterns over time changing.
The motivation for temporal contrast is that the relatively recent events are more consistent with the current state than the agelong events. 

Concerning the consistency of temporal information, the temporal-aware function $f_{t\rightarrow p}(\cdot)$ can be implemented in two opposite ways for generating the positive and negative samples: \textit{chronological probability function}~$f_{t\rightarrow p}^{tp}(\cdot)$ and \textit{reverse chronological probability function}~$f_{t\rightarrow p}^{tn}(\cdot)$. The probabilities are defined as follows:

\textbf{Chronological Probability} is proportional to the time interval between the event time $t_{u}$ and $t$: 
\begin{gather}\label{tsp}
    \hat{t}_u =  \frac{t_u-min~T_{i}^{t}}{t-min~T_{i}^{t}},\\
    p_{u}^{tp}=\frac{\exp\left(\hat{t}_u/\tau\right)}{\sum_{v\in \mathcal{N}_i^t}\exp\left(\hat{t}_{v}/\tau \right)},
\end{gather}
where $\tau$ is the temperature coefficient. 

\textbf{Reverse Chronological Probability} is inversely proportional to the time interval between the event time $t_{u}$ and $t$:
\begin{gather}
    p_{u}^{tn}=\frac{\exp\left(\widetilde{t}_u /\tau\right)}{\sum_{v\in \mathcal{N}_i^t}\exp\left(\widetilde{t}_v /\tau \right)},
\end{gather}
where $\widetilde{t}_u  = 1 - \hat{t}_u$ and $\hat{t}_u$ is defined in Eq. (\ref{tsp}). 

Given an interaction event $(i,j,t)$, \model first obtains the embedding $z_i^t$ of node $i$ at time $t$ with DGNN encoder.
Then, \model generates temporal positive and negative subgraphs by $\eta$-BFS sampling strategy with the chronological and reverse chronological probability, denoted as $\mathcal{TP}_{i}^{t}$ and $\mathcal{TN}_i^{t}$. 
For each node $u$ in $\mathcal{TP}_{i}^{t}$ and node $v$ in $\mathcal{TN}_{i}^{t}$, we retrieve the embeddings from the memory $\mathcal{M}$ and then pool them into a subgraph-wise embedding with a readout function:
\begin{align}\label{eq:dfs_readout}
    &\mathbf{h}_{tp}^{t} = \mathrm{Readout}\left(\mathbf{s}_{u}^t, u \in \mathcal{TP}_{i}^{t}\right), \\
    &\mathbf{h}_{tn}^{t} = \mathrm{Readout}\left(\mathbf{s}_{v}^t, v \in \mathcal{TN}_{i}^{t}\right),
\end{align}
where $\mathrm{Readout}(\cdot)$ is a kind of graph pooling operation \cite{ying2018hierarchical}, such as min, max, and weighted pooling. In this paper, we use mean pooling for simplicity.

The temporal contrastive learning is implemented with a triplet margin  loss~\cite{balntas2016learning} among center node embedding $z^{t}_{i}$, positive temporal subgraph embedding $\mathbf{h}_{tp}^{t}$ and negative temporal subgraph embedding $\mathbf{h}_{tn}^{t}$:
\begin{equation}\label{equ:bfs-loss}
    \gL_{\eta} = \frac{1}{|\gV^t|}\sum_{i\in\gV^t}\mathbb{E}\left\{\max\{d\left(\mathbf{z}_{i}^t,\mathbf{h}_{tp}^{t}\right)-d\left(\mathbf{z}_{i}^t,\mathbf{h}_{tn}^{t}\right)+\alpha, 0\}\right\},
\end{equation}
where $\alpha$ is the margin constant, $d(\cdot)$ is the distance metric and we adopt Euclidean distance in this paper.

Note that the long-term stable patterns are captured by the memory module of the DGNN encoder. The temporal contrast and states from the memory module in DGNN collaboratively capture the long-short term evolution patterns.

\subsubsection{\textbf{Structural Contrast (SC)}}
Besides the informative temporal evolution patterns, discriminative structural patterns are also important for structure modeling on dynamic graphs.

Following the instance discrimination paradigm, we treat the subgraph generated by $\epsilon$-DFS sampler of node $i$ as the positive sample, and the subgraph generated by $\epsilon$-DFS sampler of another random node $i'\neq i$ as the negative sample for structural contrastive learning.

Analogously, after obtaining the embedding $z_i^t$ of node $i$ at time $t$, \model generates structural positive and negative subgraphs by S-$\epsilon$-DFS sampler, denoted as $\mathcal{SP}_{i}^{t}$ and $\mathcal{SN}_{i'}^{t}$.
For each node $u$ in $\mathcal{SP}_{i}^{t}$ and node $v$ in $\mathcal{SN}_{i}^{t}$, we retrieve the embeddings from the memory $\mathcal{M}$ and then pool them into a subgraph-wise embedding with a readout function:
\begin{align}\label{equ:sp}
    &\mathbf{h}_{sp}^{t} = \mathrm{Readout}\left(\mathbf{s}_{u}^t, u \in \mathcal{SP}_{i}^{t}\right), \\
    &\mathbf{h}_{sn}^{t} = \mathrm{Readout}\left(\mathbf{s}_{v}^t, v \in \mathcal{SN}_{i'}^{t}\right).
\end{align}

The structural contrastive learning is implemented with a triplet margin loss among center node embedding $z^{t}_{i}$, positive structural subgraph embedding $\mathbf{h}_{sp}^{t}$ and negative structural subgraph embedding $\mathbf{h}_{sn}^{t}$:
\begin{equation}\label{equ:dfs-loss}
    \gL_{\epsilon} = \frac{1}{|\gV^t|}\sum_{i\in\gV^t}\mathbb{E}\left\{\max\{d\left(\mathbf{z}_{i}^t,\mathbf{h}_{sp}^{t}\right)-d\left(\mathbf{z}_{i}^t,\mathbf{h}_{sn}^{t}\right)+\alpha, 0\}\right\},
\end{equation}
where $d(\cdot)$ is the same as Eq.(\ref{equ:bfs-loss}) in temporal contrast.

\subsubsection{\textbf{Overall Pre-Training Objective Function}}
Following the general setting of unsupervised dynamic graph representation learning \cite{xu2020inductive,rossi2020temporal}, \model finally adds an auxiliary temporal link prediction pre-text task for pre-training.
Given a pair of temporal edge $(i,j,t)$, the affinity score is calculated by: 
\begin{equation}
    \hat{y}_{i,j}^t = \sigma\left(\mathrm{MLP}(\mathbf{z}_{i}^t\|\mathbf{z}_{j}^t)\right),
\end{equation}
where $\sigma(\cdot)$ denotes the sigmoid function. 
The temporal link prediction task is optimized by a binary cross-entropy loss~\cite{ruby2020binary} as follows:
\begin{equation}\label{equ:aux-loss}
    \gL_{tlp}=\sum_{(i,j,j',t)\in \gO}-y_{i,j}^t\cdot \log(\hat{y}_{i,j}^t)+y_{i,j'}^t\cdot \log(\hat{y}_{i,j'}^t),
\end{equation}
where $\gO=\{(i,j,j',t)|(i,j,t)\in \gE^t, (i,j',t)\notin \gE^t\}$, and $y_{i,j}^t\in\{0,1\}$ donates whether there is an edge between node $i$ and $j$ at time $t$. 

Overall, the objective function of \model method can be formulated as a linear combination of $\gL_{\eta}$, $\gL_{\epsilon}$ and $\gL_{tlp}$:
\begin{equation}\label{equ:beta}
    \gL_{pre} = (1-\beta)\cdot\gL_{\eta}+\beta\cdot\gL_{\epsilon}+\gL_{tlp}.
\end{equation}

where $\beta \in (0, 1)$ is a hyperparameter to balance the impact of temporal contrast and structure contrast. The \model method with any DGNN encoders is pre-trained using the gradient descent on the objective function $\gL_{pre}$.

\subsection{Evolution Information Enhanced Fine-Tuning}\label{EIE}

In the \model pre-training, the memory $\mathcal{M}$ stores the long-short term evolution information of each node through the designed pre-training objections, which is beneficial for the downstream dynamic graph tasks.
For example, if node $i$ occurs in both the pre-training and fine-tuning stages, the evolution pattern of node $i$ can be propagated to the neighborhoods in the downstream task for further enhancement.

With the above findings, we further design this evolution information enhanced fine-tuning (short as EIE) module as an optional auxiliary scheme for downstream fine-tuning.
During the pre-training of \model, we uniformly store $L$ checkpoints of the memory $\mathcal{M}$ and fuse the sequence of checkpoints as the evolution information:
\begin{equation}
\mathbf{EI}=f_{EI}\left([\mathbf{S}^{1}, ..., \mathbf{S}^{l}, ..., \mathbf{S}^{L}]\right),
\label{eq:ei}
\end{equation}
where $\mathbf{S}^{l}$ denotes $l$-th memory checkpoint from pre-training, $f_{EI}(\cdot)$ can be any kind of sequence operation function, such as mean pooling, attention mechanism, and GRU~\cite{chung2014empirical}.

In the fine-tuning stage, we first transform the evolution information $\mathbf{EI}$ with a two-layer MLP to let it better fit to the downstream data, then we combine it with the downstream temporal embeddings $\mathbf{Z}^{\mathrm{down}}$ as the enhanced embeddings:
\begin{equation}
    \mathbf{Z}^{EIE} = [\mathbf{Z}^{\mathrm{down}} \ \| \ \mathrm{MLP}(\mathbf{EI})],
\label{eq:eie}
\end{equation}
where $[\ \cdot\ \|\ \cdot\ ]$ is the matrix concatenation operator. The fine-tuning in downstream tasks is then conducted on the enhanced embeddings $\mathbf{Z}^{EIE}$.

\subsection{Model Analysis}
\subsubsection{\textbf{Pre-training stage}}
We define the time complexity of DGNN as $O(D)$, where $O(D)$ varies from different DGNN backbones. To pre-train the model scalably, the Monte Carlo trick~\cite{burch1993monte} is widely utilized for DGNNs for model training under batch processing, which is also suitable for \model. 
For \model, the additional time consumption depends on sampling and contrastive pre-training.
For sampling temporal contrastive subgraph pair of $N$ nodes, the time complexity is $O(2\eta^k N)$, where $\eta$ and $k$ denote sampling width and depth respectively.
For sampling structural contrastive subgraph pair of $N$ nodes, the time complexity is $O(2\epsilon^k N)$, where $\epsilon$ and $k$ also denote sampling width and depth.
We set $\eta=10$, $\epsilon=10$, and $k=2$ as the default setup so that $\eta^k \ll N$ and $\epsilon^k \ll N$. Then, the contrastive learning equipped with a non-parameter mean-pooling operation has the time complexity of $O(4N)$.
Therefore, the overall time complexity of \model pre-training is $O(D+2(\eta^k+\epsilon^k+2)N)$.
The pseudocode of the pre-training procedure of \model is described as the given Algorithm \ref{alg:pretrain}.

\subsubsection{\textbf{Fine-tuning stage}}
If we fine-tune \model without the EIE strategy (i.e. full fine-tune), the complexity is the same as only the DGNN backbone with $O(D)$. Then, if we fine-tune \model with EIE enhancement, the complexity comes from Eq.(\ref{eq:ei}) and Eq.(\ref{eq:eie}), of which the complexity of Eq.(\ref{eq:ei}) depends on the selected operation function of $f_{EI}(\cdot)$, and the complexity of Eq.(\ref{eq:eie}) is $O(N)$. 
Therefore, the complexity of optional fine-tuning strategies can be summarized as Table \ref{tab:finetune_comp}, where $L$ is the pre-trained checkpoint length and we set it as 10 for default. The performance comparison results with different fine-tuning ways are also conducted in the following Subsection \ref{exp:eie}.
The procedure of evolution information enhanced fine-tuning in \model is illustrated at Algorithm \ref{alg:finetune}.

\begin{algorithm}[htbp]
    \renewcommand{\algorithmicrequire}{\textbf{Input:}}  
    \renewcommand{\algorithmicensure}{\textbf{Output:}} 
    \caption{The Pre-Training Procedure of \model}  
    \label{alg:pretrain}
    \begin{algorithmic}[1]
        \REQUIRE Dynamic graph $\gG=(\gV^T, \gE^T)$; $\eta$-BFS sampler $f_{\eta}(\cdot)$; $\epsilon$-DFS sampler $f_{\epsilon}(\cdot)$; Hyperparameters $\beta$; Pre-training epochs $P$.
        \ENSURE Pre-trained DGNN model parameters $\theta^*$ for downstream fine-tuning.
        \STATE Randomly initialize the DGNN model $f_{\theta}$ and the trainable parameters in \model.
        \FOR{$epoch \in 1,2,...P$}
        \FORALL{$(i, j, t) \in \gG$ sorted by timestamp}
        \STATE Obtain temporal contrast sample pair $\mathcal{TP}_{i}^{t}$ and $\mathcal{TN}_{i}^{t}$ by $f_{\eta}(\cdot)$;
        \STATE Obtain structural contrast sample pair $\mathcal{SP}_{i}^{t}$ and $\mathcal{SN}_{i}^{t}$ by $f_{\epsilon}(\cdot)$;
        \STATE Conduct the temporal contrast learning with the sampled $\mathcal{TP}_{i}^{t}$ and $\mathcal{TN}_{i}^{t}$, and calculate the temporal contrast loss $\gL_{\eta}$ via Eq.(\ref{equ:bfs-loss});
        \STATE Conduct the structural contrast learning with the sampled $\mathcal{SP}_{i}^{t}$ and $\mathcal{SN}_{i}^{t}$, and calculate the temporal contrast loss $\gL_{\epsilon}$ via Eq.(\ref{equ:dfs-loss});
        \STATE Conduct the auxiliary pre-text task and calculate the corresponding loss $\gL_{tlp}$ via Eq.(\ref{equ:aux-loss});
        \ENDFOR
        \STATE Minimize the joint pre-training loss $\gL_{pre}=(1-\beta)\cdot \gL_{\eta} + \beta \cdot \gL_{\epsilon} + \gL_{tlp}$;
        \STATE Update model's learnable parameters by using stochastic gradient descent;
        \STATE \textit{// Optional for fine-tuning}
        \STATE Save the checkpoint from the memory $\mathcal{M}$ in $f_{\theta}$;
        \ENDFOR
        \STATE Save the pre-trained DGNN parameters $\theta^*$.
    \end{algorithmic}  
\end{algorithm}

\begin{table}[htbp]
  \centering
  \caption{Complexity analysis of different fine-tuning ways.}
    \begin{tabular}{c|c}
    \toprule
    Fine-tuning Strategy & Complexity \\
    \midrule
    \model-(w/o EIE) &  $O(D)$ \\
    \midrule
    \model-(EIE-mean) &  $O(D+N+1)$ \\
    \model-(EIE-attn) &  $O(D+2N)$ \\
    \model-(EIE-GRU) &  $O(D+N+NL^2)$ \\
    \bottomrule
    \end{tabular}%
  \label{tab:finetune_comp}%
\end{table}%


\begin{algorithm}[htbp]
    \renewcommand{\algorithmicrequire}{\textbf{Input:}}  
    \renewcommand{\algorithmicensure}{\textbf{Output:}} 
    \caption{The Evolution Information Enhanced Fine-Tuning Procedure of \model}  
    \label{alg:finetune}
    \begin{algorithmic}[1]
        \REQUIRE Pre-trained DGNN model $f_{\theta^*}$; Dynamic graph $\gG=(\gV^T, \gE^T)$; Sequence of evolution checkpoints $[\mathbf{S}^{1}, ..., \mathbf{S}^{L}]$; Evolution information fusion function $f_{EI}(\cdot)$; Fine-tuning epochs $F$.
        \ENSURE Model performance on the downstream task.
        \FOR{$epoch \in 1,2,...F$}
        \STATE Obtain downstream node embeddings $\mathbf{Z}^{down}$ by the pre-trained DGNN model $f_{\theta^*}$;
        \STATE Fuse $[\mathbf{S}^{1}, ..., \mathbf{S}^{L}]$ by $f_{EI}(\cdot)$ and obtain the evolution information $\mathbf{EI}$;
        \STATE Obtain the enhanced embeddings $\mathbf{Z}^{EIE}$ with $\mathbf{Z}^{down}$ and $\mathbf{EI}$ via Eq.(\ref{eq:eie});
        \STATE Conduct the downstream task with $\mathbf{Z}^{EIE}$ and obtain the corresponding fine-tuning loss $\gL_{down}$;
        \STATE Minimize $\gL_{down}$ and fine-tune DGNN's parameters.
        \ENDFOR
    \end{algorithmic}  
\end{algorithm}

%% file: experiment_new.tex

In this section, we conduct extensive experiments on five public widely-used dynamic graph datasets and an industrial dynamic graph dataset from Meituan to demonstrate the effectiveness of the proposed \model method from various perspectives.
To evaluate the generalization ability of \model, we consider three \textit{transfer settings} and two \textit{downstream tasks}. 1) Transfer settings include time transfer, field transfer, and time+field transfer. 2) Downstream tasks include both \textit{dynamic link prediction} and \textit{dynamic node classification}.

\subsection{Experimental Datasets}
The experiments are conducted on five widely-used research datasets and an industrial dataset from Meituan on various downstream tasks.
The details of these datasets are introduced as follows.

\noindent\textit{\textbf{Dynamic Link Prediction Datasets:}}

\textbf{Amazon Review}\footnote{\url{https://jmcauley.ucsd.edu/data/amazon}}~\cite{ni2019justifying} is a large-scale user-product dynamic graph dataset containing 20.9 million users, 9.3 million products, and 82.8 million reviews (edges) into 29 product fields. The time span of these reviews is in the range of May 1996 - Oct 2018.
We select the \textit{Beauty}, \textit{Luxury}, and \textit{Arts, Crafts, and Sewing} fields for the experiment.
The \textit{user-product interaction prediction} is set as the downstream dynamic link prediction task for Amazon dataset.

\textbf{Gowalla}\footnote{\url{http://www.yongliu.org/datasets.html}}~\cite{liu2014exploiting} is a famous social network for users check-in at various locations, containing about 36 million check-ins made by 0.32 million users over 2.8 million locations. These check-in records are in the time span of Jan 2009 - June 2011. The locations are grouped into 7 main fields. We select the \textit{Entertainment}, \textit{Outdoors}, and \textit{Food} fields for the experiment.
The \textit{check-in prediction} is set as the downstream dynamic link prediction task for Gowalla dataset.

The data split of Amazon and Gowalla varies from different transfer settings considering both time and field, which is detailed described in Subsection \ref{sec:setup}.

\textbf{Meituan} is a \textit{industrial} user-poi interaction (i.e. click and purchase) dataset collected from Meituan food delivery service in Beijing of China from Feb.14th to Mar.28th, 2022 (42 days), which contains about 0.75 million interaction records (sorted by time).
Each interaction record contains the user-ID, poi-ID, timestamp, and other contexts.
We utilize this industrial dataset for evaluating the dynamic link prediction task with a ratio of 6:4 sorted by time for pre-training and downstream.

\noindent\textit{\textbf{Dynamic Node Classification Datasets:}}

\textbf{Wikipedia}\footnote{\url{http://snap.stanford.edu/jodie/wikipedia.csv}}~\cite{kumar2018learning} is a dynamic network between users and edited pages for node classification. It contains 9,227 nodes and around 157,474 temporal edges. Dynamic labels indicate banned users.

\textbf{MOOC}\footnote{\url{http://snap.stanford.edu/jodie/mooc.csv}}~\cite{kumar2018learning} is a dynamic network of students and online courses for node classification, which contains 7,145 nodes and 411,749 interactions between them. Dynamic labels indicate drop-out students.

\textbf{Reddit}\footnote{\url{http://snap.stanford.edu/jodie/reddit.csv}}~\cite{kumar2018learning} is a dynamic graph between active users and their posts under subreddits for node classification, with about 11,000 nodes and 672,447 temporal edges, and dynamic labels indicating banned users.

We split each dynamic node classification dataset with the ratio of 6:2:1:1 sorted by time for pre-training, downstream training, validation, and testing, respectively.

\begin{table*}[htbp]
  \centering
  \vspace{-0.6em}
  \caption{Detailed statistics of the experimental datasets with both time and field information. (T: Time Transfer, F: Field Transfer, T+F: Time+Field Transfer).}
  \resizebox{\linewidth}{!}{
    \begin{tabular}{c|c|c|c|r|r|r|c|c|r|r|r}
    \toprule
    \multicolumn{2}{c|}{Dataset} & \multicolumn{5}{c|}{Amazon}           & \multicolumn{5}{c}{Gowalla} \\
    \hline
    \multicolumn{2}{c|}{Data Description} & Data Field & Data Range & \# Nodes & \# Edges & \multicolumn{1}{c|}{Density} & Data Field & Data Range & \# Nodes & \# Edges & \multicolumn{1}{c}{Density} \\
    \hline
    \multirow{4}{*}{Pre-train} & \multirow{2}{*}{T} & Beauty & May 1996-Dec 2016 &   261,889    & 270,714  &   0.0008\%    & Entertainment & Jan 2009-Dec 2010 &   208,309    &    223,777   & 0.0010\% \\
 \cline{3-12}
&       & Luxury & May 1996-Dec 2016 &  303,542     &    401,987   & 0.0009\%   & Outdoors & Jan 2009-Dec 2010 &    223,777   &     1,192,387  & 0.0047\% \\
 \cline{2-12}
& F     & Arts, Crafts, and Sewing & Jan 2017-Oct 2018 &   729,282    &   897,079    &     0.0003\%  & Food  & Jan 2011-June 2011 &   673,858    &    2,708,688   & 0.0011\% \\
 \cline{2-12}
& T+F   & Arts, Crafts, and Sewing & May 1996-Dec 2016 &   1,345,428    &   1,978,838    &      0.0002\% & Food  & Jan 2009-Dec 2010 &   679,586    &  5,067,244    & 0.0021\% \\
    \hline
    \multirow{2}{*}{Downstream} & \multirow{2}{*}{T/F/T+F} & Beauty & Jan 2017-Oct 2018 &   107,916    &   100,631   &   0.0017\%    & Entertainment & Jan 2011-June 2011 &   199,547    &    733,863   & 0.0036\% \\
 \cline{3-12}
&       & Luxury & Jan 2017-Oct 2018 &   143,686    &  172,641     &    0.0013\%   & Outdoors & Jan 2011-June 2011 &  227,980     &     763,764  & 0.0029\% \\
    \bottomrule
    \end{tabular}%
    }
    \vspace{-0.8em}
  \label{tab:statistic}%
\end{table*}%

\begin{table}[htbp]
  \centering
  \vspace{-0.6em}
  \caption{Detailed statistics of the datasets without field information.}
  \resizebox{0.835\linewidth}{!}{
    \begin{tabular}{l|cccc}
    \toprule
    Dataset & \# Nodes & \# Edges & \# Timespan & Density \\
    \midrule
    Meituan & $\sim$45,000 & $\sim$750,000 & 42 days & 0.0741\% \\
    \midrule
    Wikipedia & 9,227 & 157,474 & 30 days & 0.3700\% \\
    MOOC  & 7,145 & 411,749 & 30 days &  1.6133\% \\
    Reddit & 11,000 & 672,447 & 30 days &  1.1116\% \\
    \bottomrule
    \end{tabular}%
    }
    \vspace{-0.8em}
  \label{tab:da2}%
\end{table}%

\subsection{Baselines}
We compare the proposed \model method with ten representative methods, including state-of-the-art static graph learning methods and dynamic graph learning methods, which can be organized into four main categories:

\noindent \textit{\textbf{Task-supervised Static Graph Learning Models:}}
\begin{itemize}[leftmargin=*]
    \item \textbf{GraphSAGE}~\cite{hamilton2017inductive} learns node representation by sampling and aggregating features from the node’s local neighborhood without temporal information. It performs link prediction as its pre-training task.
    \item \textbf{GAT}~\cite{velivckovic2018graph} aggregates the neighborhood message with the multi-head attention mechanism. The pre-training task of GAT is the same as GraphSAGE.
    \item \textbf{GIN}~\cite{xu2018powerful} is a simple GNN architecture that generalizes the Weisfeiler-Lehman test. It also performs the link prediction task for the model pre-training.
\end{itemize}

\noindent \textit{\textbf{Self-supervised Static Graph Learning Models:}}
\begin{itemize}[leftmargin=*]
    \item \textbf{DGI}~\cite{velickovic2019deep} maximizes the mutual information between local patch representation and corresponding global graph summary in a self-supervised manner.
    \item \textbf{GPT-GNN}~\cite{hu2020gpt} is a state-of-the-art generative pre-training framework with self-supervised node attribute generation and edge generation tasks on large-scale graph pre-training.
\end{itemize}

\noindent \textit{\textbf{Task-supervised Dynamic Graph Learning Models:}}
\begin{itemize}[leftmargin=*]
    \item \textbf{DyRep}~\cite{trivedi2019dyrep} is a temporal point process (TPP-based) dynamic graph model, which posits representation learning as a latent mediation process. Following its task setting for dynamic graphs, we adopt temporal link prediction as its pre-training task.
    \item \textbf{JODIE}~\cite{kumar2018learning} is a state-of-the-art dynamic graph model, which employs two recurrent neural networks and a novel projection operator to estimate the embedding of a node at any time in the future. The pre-training task of JODIE is the same as DyRep. 
    \item \textbf{TGN}~\cite{rossi2020temporal} is a generic and state-of-the-art dynamic graph learning framework with memory modules and dynamic graph-based operators. The pre-training task of TGN is also the same as DyRep. 
\end{itemize}

\noindent \textit{\textbf{Self-supervised Dynamic Graph Learning Models:}}
\begin{itemize}[leftmargin=*]
    \item \textbf{DDGCL}~\cite{tian2021self} is a self-supervised dynamic graph learning method via contrasting two nearby temporal views of the same node identity, with a time-dependent similarity critic and GAN-type contrastive loss.
    \item \textbf{SelfRGNN}~\cite{sun2022self} is a state-of-the-art self-supervised Riemannian dynamic graph neural network with the Riemannian reweighting self-contrast for dynamic graph learning.
\end{itemize}

For all the baselines, we first pre-train them on the same pre-training data as \model and utilize them for initialization with full fine-tuning strategy \cite{qiu2020gcc} in downstream tasks. 

\subsection{Experimental Setup} \label{sec:setup}
To evaluate the effectiveness of the \model method, we follow the experimental setting of popular existing pre-training work \cite{hu2020gpt} with three \textit{transfer settings} and two \textit{downstream tasks} in fine-tuning on these datasets. 

\noindent \textit{\textbf{Transfer Settings:}}
\begin{itemize}[leftmargin=*]
    \item \textbf{Time transfer}. We split the graph by time spans for pre-training and fine-tuning on the same field. 
    For the Amazon Review dataset, we split the data before \textit{2017} for pre-training and data since \textit{2017} for fine-tuning. 
    For the Gowalla dataset, we split the data before \textit{2011} for pre-training and data since \textit{2011} for fine-tuning.
    For Meituan, Wikipedia, MOOC, and Reddit datasets, due to the lack of field categories, we only conduct the \textit{time transfer} and adopt the same chronological data split with the first 60\% for pre-training and the rest for fine-tuning.
    \item \textbf{Field transfer}. We pre-train the model from one graph field and transfer it to other graph fields. For the Amazon Review dataset, we use data from \textit{Arts, Crafts, and Sewing} field for pre-training and data from \textit{Beauty} and \textit{Luxury} fields for fine-tuning. For the Gowalla dataset, we use data from the \textit{Food} field for pre-training and data from \textit{Entertainment} and \textit{Outdoors} fields for fine-tuning. 
    \item \textbf{Time+Field transfer}.
    In this setting, we consider graph transfer from different time spans and fields simultaneously.
    That is, we use the data from fields before a particular time for model pre-training and the data from other fields after a particular time for fine-tuning. The split time and fields are the same as the above two tasks: For the Amazon dataset, we first pre-train the model on \textit{Arts, Crafts, and Sewing} field under the data range from May 1996 to Dec 2016, and then we fine-tune the DGNN model on \textit{Beauty} or \textit{Luxury} data fields under the data range from Jan 2017 to Oct 2018. For the Gowalla dataset, we pre-train the model on \textit{Food} field under the data range from Jan 2009 to Dec 2010, then we fine-tune the DGNN model on \textit{Entertainment} or \textit{Outdoors} fields under the data range from Jan 2011 to June 2011.
\end{itemize}

\noindent \textit{\textbf{Downstream Task Settings:}}
\begin{itemize}[leftmargin=*]
    \item \textbf{Dynamic link prediction}. For Amazon Review, Gowalla, and Meituan datasets, we aim to predict future user-item interactions at a specific time as the downstream task. 
    Then, we adopt the widely used AUC and AP metrics as the evaluation metrics. 
    Note that we conduct the downstream task in \textit{Beauty} and \textit{Luxury} fields for the Amazon Review dataset, and in \textit{Entertainment} and \textit{Outdoors} fields for Gowalla.
    \item \textbf{Dynamic node classification}. For Wikipedia, MOOC, and Reddit datasets, we aim to predict the node state with a specific class label at a specific time as the downstream task.
    Then, we adopt the widely used AUC metric as the node classification evaluation metric.
\end{itemize}

We introduce the statistics and settings of the experimental datasets in Table \ref{tab:statistic} and Table \ref{tab:da2}.

\noindent \textit{\textbf{Environment Configuration and Hyper-parameter Setting:}}

The model hyper-parameters are optimized via grid search on all the experimental datasets, and the best models are selected by early stopping based on the AUC score on the validation set. Hyper-parameter ranges of the model for grid search are the following: learning rate in $\{5 \times 10^{-4}, 1 \times 10^{-4}, 5 \times 10^{-3}, 1 \times 10^{-3}, 5 \times 10^{-2}, 1 \times 10^{-2}\}$, the batch size settings are searched from range $\{256, 512, 1024, 2048, 4096, 8192\}$ for the link prediction task due to large-scale property, and the batch size of the node classification task is set to 512. All weighting matrices are initialized by Xavier initialization for all models. For all dynamic graphs (Dyrep, JODIE, TGN, DDGCL, SelfRGNN, \model), the memory states are all initialized with zero.
Note that we run all the experiments \textit{five} times with different random seeds and report the average results with standard deviation to prevent extreme cases.
All experiments are conducted on NVIDIA A100 (80G) GPUs.


\begin{table*}[htbp]
  \centering
  \vspace{-0.6em}
  \caption{Comparison results on dynamic link prediction task. The best and second-best results in each column are highlighted in \textbf{bold} font and \underline{underlined}. NaN denotes not a number of the value of loss function during the pre-training stage.}
  \resizebox{\linewidth}{!}{
    \begin{tabular}{cc|cc|cc|cc|cc}
    \toprule
    \multicolumn{2}{c|}{Downstream Dataset} & \multicolumn{4}{c|}{Amazon}   & \multicolumn{4}{c}{Gowalla} \\
    \midrule
    \multicolumn{2}{c|}{Evaluation Task} & \multicolumn{2}{c|}{Beauty} & \multicolumn{2}{c|}{Luxury} & \multicolumn{2}{c|}{Entertainment} & \multicolumn{2}{c}{Outdoors} \\
    \midrule
    \multicolumn{2}{c|}{Evaluation Metric} & AUC   & AP    & AUC   & AP    & AUC   & AP    & AUC   & AP \\
    \midrule
    \multirow{13}[10]{*}{\rotatebox{90}{Time Transfer}} & GraphSAGE & 0.7537±0.0062 & 0.6896±0.0053 & 0.6395±0.0040 & 0.5814±0.0028 & 0.6315±0.0018 & 0.6906±0.0009 & 0.6183±0.0007 & 0.6797±0.0003 \\
          & GIN   & 0.6908±0.0100 & 0.6904±0.0073 & 0.5948±0.0014 & 0.5815±0.0014 & 0.5179±0.0015 & 0.6189±0.0007 & 0.5154±0.0021 & 0.6213±0.0016 \\
          & GAT   & 0.5217±0.0015 & 0.6123±0.0023 & 0.5403±0.0012 & 0.6292±0.0020 & 0.5315±0.0005 & 0.6242±0.0007 & 0.5420±0.0032 & 0.6334±0.0024 \\
\cmidrule{2-10}          & DGI   & 0.6928±0.0047 & 0.6773±0.0018 & 0.6083±0.0017 & 0.5805±0.0024 & 0.5763±0.0037 & 0.6541±0.0025 & 0.5955±0.0039 & 0.6675±0.0012 \\
          & GPT-GNN & 0.5785±0.0010 & 0.5650±0.0018 & 0.5532±0.0008 & 0.5267±0.0008 & 0.5139±0.0019 & 0.5157±0.0017 & 0.5118±0.0022 & 0.5090±0.0018 \\
\cmidrule{2-10}          & DyRep & 0.8023±0.0073 & 0.8152±0.0020 & 0.7853±0.0023 & 0.7906±0.0029 & 0.8490±0.0124 & 0.8586±0.0093 & 0.8269±0.0044 & 0.8337±0.0057 \\
          & JODIE & 0.8472±0.0149 & 0.8411±0.0512 & \underline{0.8201}±0.0030 & 0.8147±0.0021 & 0.8572±0.0035 & 0.8669±0.0021 & 0.8274±0.0034 & 0.8390±0.0040 \\
          & TGN   & 0.8589±0.0025 & 0.8533±0.0016 & 0.7985±0.0049 & 0.7843±0.0049 & \underline{0.9152}±0.0032 & \underline{0.9076}±0.0058 & \underline{0.9051}±0.0046 & \underline{0.9073}±0.0035 \\
\cmidrule{2-10}          & DDGCL & 0.8146±0.0025 & 0.8474±0.0023 & 0.8066±0.0007 & \underline{0.8153}±0.0015 & 0.7117±0.0012 & 0.7586±0.0012 & 0.6617±0.0017 & 0.7087±0.0014 \\
          & SelfRGNN & 0.6352±0.0253 & 0.6011±0.0312 & 0.5744±0.0405 & 0.5576±0.0371 & 0.5457±0.0113 & 0.5316±0.0107 & 0.5467±0.0192 & 0.5332±0.0164 \\
\cmidrule{2-10}          & {\textbf{CPDG} (DyRep)}  & {0.8275±0.0028} & {0.8371±0.0020} & {0.7976±0.0031} & {0.8038±0.0037} & {0.8883±0.0040} & {0.8946±0.0043} & {0.8459±0.0020} & {0.8575±0.0001} \\
& {\textbf{CPDG} (JODIE)} & \underline{0.8672}±0.0017 & \textbf{0.8665}±0.0017 & \textbf{0.8378}±0.0034 & \textbf{0.8390}±0.0041 & {0.8914±0.0010} & {0.8966±0.0004} & {0.8709±0.0004} & {0.8767±0.0008} \\
& {\textbf{CPDG} (TGN)}  & {\textbf{0.8690}±0.0026} & {\underline{0.8594}±0.0022} & {0.8042±0.0068} & {0.7931±0.0048} & {\textbf{0.9234}±0.0019} & {\textbf{0.9202}±0.0042} & {\textbf{0.9134}±0.0026} & {\textbf{0.9121}±0.0045} \\
    \midrule
    \midrule
    \multirow{13}[10]{*}{\rotatebox{90}{Field Transfer}} & GraphSAGE & 0.7265±0.0232 & 0.6755±0.0145 & 0.6166±0.0067 & 0.5807±0.0046 & 0.6330±0.0012 & 0.6875±0.0006 & 0.6284±0.0025 & 0.6829±0.0012 \\
          & GIN   & 0.6652±0.0092 & 0.6516±0.0036 & 0.5782±0.0053 & 0.5651±0.0072 & 0.5167±0.0005 & 0.5886±0.0008 & 0.5176±0.0043 & 0.5598±0.0021 \\
          & GAT   & 0.5161±0.0120 & 0.5585±0.0243 & 0.5635±0.0380 & 0.5540±0.0195 & 0.5332±0.0004 & 0.6112±0.0005 & 0.5312±0.0006 & 0.6035±0.0011 \\
\cmidrule{2-10}          & DGI   & 0.6922±0.0051 & 0.6408±0.0030 & 0.6027±0.0028 & 0.5731±0.0024 & 0.5724±0.0014 & 0.6479±0.0024 & 0.5849±0.0002 & 0.6505±0.0002 \\
          & GPT-GNN & 0.5777±0.0013 & 0.5664±0.0020 & 0.5528±0.0006 & 0.5272±0.0009 & 0.5136±0.0017 & 0.5098±0.0011 & 0.5106±0.0016 & 0.5040±0.0009 \\
\cmidrule{2-10}          & DyRep & 0.8054±0.0023 & 0.7946±0.0073 & 0.7788±0.0047 & 0.7894±0.0098 & 0.8589±0.0077 & 0.8658±0.0057 & 0.8395±0.0057 & 0.8467±0.0063 \\
          & JODIE & 0.8121±0.0027 & 0.8074±0.0042 & 0.7812±0.0039 & 0.7878±0.0048 & 0.8495±0.0090 & 0.8587±0.0075 & 0.8409±0.0027 & 0.8483±0.0030 \\
          & TGN   & \underline{0.8391}±0.0052 & \underline{0.8347}±0.0033 & 0.7753±0.0039 & 0.7628±0.0039 & \textbf{0.8877}±0.0050 & \underline{0.8788}±0.0061 & \underline{0.8787}±0.0033 & \underline{0.8725}±0.0054 \\
\cmidrule{2-10}          & DDGCL & 0.7929±0.0021 & 0.8300±0.0023 & \underline{0.7854}±0.0012 & \underline{0.8014}±0.0011 & 0.7202±0.0009 & 0.7672±0.0007 & 0.6721±0.0008 & 0.7196±0.0009 \\
          & SelfRGNN & 0.5313±0.0040 & 0.5315±0.0067 & 0.5140±0.0041 & 0.5110±0.0002 & 0.5051±0.0021 & 0.5097±0.0038 & 0.5123±0.0062 & 0.5106±0.0045 \\
\cmidrule{2-10}          & {\textbf{CPDG} (DyRep)}  & {0.8124±0.0012} & {0.8226±0.0018} & {0.7827±0.0016} & {0.8008±0.0022} & {0.8659±0.0049} & {0.8742±0.0036} & {0.8512±0.0014} & {0.8543±0.0023} \\
    & {\textbf{CPDG} (JODIE)} & 0.8220±0.0148 & 0.8255±0.0121 & \textbf{0.8296}±0.0032 & \textbf{0.8299}±0.0017 & {0.8734±0.0012} & {0.8781±0.0014} & {0.8516±0.0015} & {0.8573±0.0019} \\
    & {\textbf{CPDG} (TGN)} & {\textbf{0.8439}±0.0040} & {\textbf{0.8372}±0.0038} & {0.7782±0.0030}
 & {0.7641±0.0065} & \underline{0.8870}±0.0063 & \textbf{0.8800}±0.0052 & \textbf{0.8868}±0.0048 & \textbf{0.8841}±0.0041 \\
    \midrule
    \midrule
    \multirow{13}[10]{*}{\rotatebox{90}{Time+Field Transfer}} & GraphSAGE & 0.7428±0.0047 & 0.6730±0.0034 & 0.6296±0.0047 & 0.5776±0.0044 & 0.5118±0.0141 & 0.5102±0.0090 & 0.5051±0.0109 & 0.5059±0.0120 \\
          & GIN   & 0.6696±0.0150 & 0.6543±0.0062 & 0.5854±0.0070 & 0.5741±0.0119 & 0.5089±0.0009 & 0.6053±0.0010 & 0.5111±0.0012 & 0.6089±0.0006 \\
          & GAT   & 0.5206±0.0115 & 0.5796±0.0080 & 0.5268±0.0027 & 0.5422±0.0153 & 0.5291±0.0005 & 0.6141±0.0014 & 0.5403±0.0007 & 0.6168±0.0012 \\
\cmidrule{2-10}          & DGI   & 0.6846±0.0021 & 0.6524±0.0034 & 0.5990±0.0038 & 0.5757±0.0037 & 0.5714±0.0018 & 0.6374±0.0026 & 0.5843±0.0046 & 0.6525±0.0039 \\
          & GPT-GNN & 0.5773±0.0016 & 0.5665±0.0006 & 0.5531±0.0005 & 0.5271±0.0008 & 0.5105±0.0020 & 0.5074±0.0022 & 0.5098±0.0030 & 0.5029±0.0029 \\
\cmidrule{2-10}          & DyRep & 0.8026±0.0087 & 0.8001±0.0067 & 0.7726±0.0064 & 0.7795±0.0074 & 0.8458±0.0113 & 0.8548±0.0092 & 0.8250±0.0038 & 0.8325±0.0041 \\
          & JODIE & 0.8401±0.0199 & 0.8346±0.0165 & \underline{0.8115}±0.0114 & \underline{0.8124}±0.0094 & 0.8412±0.0080 & 0.8538±0.0069 & 0.8272±0.0063 & 0.8372±0.0059 \\
          & TGN   & 0.8478±0.0079 & 0.8449±0.0065 & 0.7820±0.0069 & 0.7739±0.0071 & 0.8622±0.0040 & 0.8607±0.0049 & \underline{0.8596}±0.0078 & \underline{0.8527}±0.0105 \\
\cmidrule{2-10}          & DDGCL & 0.8060±0.0009 & 0.8408±0.0009 & 0.8037±0.0009 & 0.8123±0.0019 & 0.7194±0.0009 & 0.7668±0.0006 & 0.6697±0.0015 & 0.7176±0.0011 \\
          & SelfRGNN & 0.5374±0.0100 & 0.5586±0.0184 & 0.5156±0.0037 & 0.5144±0.0018 & NaN   & NaN   & NaN   & NaN \\
\cmidrule{2-10}          & {\textbf{CPDG} (DyRep)}  & {0.8113±0.0047} & {0.8115±0.0073} & {0.7746±0.0031} & {0.7901±0.0027} & {\underline{0.8679}±0.0073} & {\textbf{0.8779}±0.0056} & {0.8350±0.0106} & {0.8487±0.0090} \\
    & {\textbf{CPDG} (JODIE)} & \textbf{0.8622}±0.0031 & \textbf{0.8541}±0.0033 & \textbf{0.8250}±0.0028 & \textbf{0.8250}±0.0034 & {0.8667±0.0051} & {\underline{0.8738}±0.0027} & {0.8405±0.0110} & {0.8490±0.0083} \\
    & {\textbf{CPDG} (TGN)}  & {\underline{0.8597}±0.0031} & {\underline{0.8523}±0.0040} & {0.7896±0.0026} & {0.7766±0.0035} & \textbf{0.8732}±0.0065 & 0.8661±0.0090 & \textbf{0.8720}±0.0053 & \textbf{0.8678}±0.0044 \\
    \bottomrule
    \end{tabular}%
    }
  \label{tab:maintable}
  \vspace{-0.8em}
\end{table*}%

\subsection{Main Results} \label{exp:main}
In this section, we report the comparison results between \model and the baselines for various downstream tasks, including dynamic link prediction and node classification.
{The hyper-parameters of \model are set as $\eta=10$, $\epsilon=10$, $k=2$, and $L=10$ in the main results, and we will analyze the impact of changes in these hyper-parameters in the Subsection \ref{exp:param}.}

\noindent\textit{\textbf{Performance on Dynamic Link Prediction Task:}}

The results and comparisons on Amazon Review and Gowalla datasets under three transfer settings are shown in Table \ref{tab:maintable}. And the results and comparisons on Meituan under the time transfer setting are shown in Table \ref{tab:meituan_result}. From the results, we have the following observations.
\begin{itemize}[leftmargin=*]
    \item \textbf{Our proposed \model achieves the best performance under different transfer settings.} {Specifically, the average improvements over the best-performed baseline on the AP metric of all the fields on Amazon Review and Gowalla datasets are 1.59\%, 1.33\%, 1.60\%  w.r.t. \textit{time transfer}, \textit{field transfer}, and \textit{time+field transfer} settings, respectively.
    Furthermore, in the Meituan industrial dataset, compared to the DGNN models, the performance gains by \model can also be observed.
    These observations demonstrate the effectiveness of our \model through flexible structural-temporal samplers along with temporal-view and structural-view subgraph contrastive pre-training schemes and optional evolution information enhanced fine-tuning scheme.}
    \item \textbf{The dynamic graph methods generally perform significantly better than the static graph methods.} Among the compared baselines, the dynamic graph methods (DyRep, JODIE, TGN, DDGCL, and SelfRGNN) generally perform significantly better than the static graph methods (GraphSAGE, GIN, GAT, DGI, and GPT-GNN) on all the datasets.  
    This further verifies the necessity of capturing the temporal evolution patterns in the dynamic graphs. 
    Another interesting finding is that in dynamic graph tasks, the static generative graph pre-training framework performs relatively worse than other static GNNs, such as GraphSAGE, which has also been observed and discussed in \cite{hou2022graphmae}. 
    \item {\textbf{The dynamic task-supervised methods generally perform better than dynamic self-supervised methods.}
    Generally, among the compared baselines, the dynamic task-supervised methods (DyRep, JODIE, and TGN) perform better than the dynamic self-supervised methods (DDGCL and SelfRGNN) on experimental datasets.
    This demonstrates the importance of memory in the DGNN encoder to capture the long-term evolution of each node in dynamic graphs and the insufficiency for general pattern modeling of current dynamic self-supervised learning manners under the pre-training setting.}
\end{itemize}

\begin{table}[tbp]\label{exp:meituan}
  \centering
  \caption{Comparison results on the Meituan industry dataset.}
    \vspace{-0.6em}
  \resizebox{0.71\linewidth}{!}{
    \begin{tabular}{c|cc}
    \toprule
    Dataset & \multicolumn{2}{c}{Meituan} \\
    \midrule
    Metric & AUC   & AP \\
    \midrule
    DyRep &  0.8461±0.0028 & 0.8355±0.0023\\
    \textbf{CPDG} (DyRep) &  \textbf{0.8472}±0.0061   & \textbf{0.8372}±0.0033 \\
    \midrule
    JODIE &  0.8498±0.0042     &  0.8315±0.0077 \\
    \textbf{CPDG} (JODIE) &  \textbf{0.8513}±0.0042    &  \textbf{0.8398}±0.0034 \\
    \midrule
    TGN   &   0.8431±0.0040    &  0.8304±0.0037 \\
    \textbf{CPDG} (TGN) & \textbf{0.8480}±0.0037 &  \textbf{0.8364}±0.0047 \\
    \bottomrule
    \end{tabular}%
    }
    \vspace{-0.8em}
  \label{tab:meituan_result}%
\end{table}%

\noindent \textit{\textbf{Performance on Dynamic Node Classification Task:}}

In addition to the dynamic link prediction downstream task, we also conduct the dynamic node classification task on Wikipedia, MOOC, and Reddit datasets compared with the state-of-the-art dynamic graph methods to further verify the generalization ability in various downstream tasks for \model.
Table \ref{tab:nc_results} shows the node classification results. From these results, we have the following observation:
\begin{itemize}[leftmargin=*]
    \item \textbf{\model also has the best performance under the dynamic node classification task in most cases}, which shows \model has good generalization capability in various downstream tasks.
    Note that the performance of \model is lower than TGN on the MOOC dataset.
    One reason is that the generic structure and temporal patterns are not as obvious in the MOOC dataset as in other datasets.
    \item \textbf{The dynamic task-supervised methods make a better performance than the dynamic self-supervised methods}. 
    {These results are consistent with the observation in dynamic link prediction, which further verifies the effectiveness of the memory module within DGNNs in learning long-term evolution. And the superiority of \model with generally all the equipped backbone DGNNs also illustrates the effectiveness of the model design for capturing other generic patterns.}
\end{itemize}

\begin{table}[t]
  \centering
  \vspace{-0.6em}
  \caption{Comparison results of AUC metrics on dynamic node classification task under \textit{time+transfer} setting.}
  \resizebox{\linewidth}{!}{
    \begin{tabular}{c|ccc}
    \toprule
    Method & Wikipedia & MOOC & Reddit\\
    \midrule
    DyRep &  0.8189±0.0098     & 0.6342±0.0108 & 0.5614±0.0281\\
    JODIE &    0.8206±0.0067   &  0.6185±0.0049 & 0.5385±0.0303\\
    TGN   &   0.8302±0.0012    & \textbf{0.7009}±0.0029 &  0.5552±0.0106\\
    \midrule
    DDGCL &  0.7091±0.0162  & 0.5674±0.0061 & 0.5205±0.0207\\
    SelfRGNN &  0.8490±0.0024   &  0.6051±0.0027 & 0.5363±0.0051\\
    \midrule
    {\textbf{CPDG} (DyRep)} &  {0.8323±0.0059}   &  {0.6689±0.0014} & {\underline{0.6110}±0.0101} \\
    {\textbf{CPDG} (JODIE)} &  {\textbf{0.8649}±0.0023}   &  {0.6551±0.0019} & {0.5474±0.0089} \\
    \textbf{CPDG} (TGN) &  \underline{0.8558}±0.0024   &  \underline{0.6797}±0.0254 & \textbf{0.6551}±0.0118\\
    \bottomrule
    \end{tabular}%
    }
    \vspace{-1em}
  \label{tab:nc_results}
\end{table}%

\subsection{Model Generalization}\label{exp:general}
\label{subsec:general}

{\textbf{Backbone Generalization}. To investigate the generalization of the proposed \model method, we evaluate the performance of \model with different DGNN encoders.
We try three representative DGNN encoders namely, DyRep, JODIE, and TGN. 
More specifically, we experiment on each encoder being pre-trained with our \model and perform on all the datasets of both dynamic link prediction and dynamic node classification tasks.
The results shown in Table \ref{tab:maintable}, Table \ref{tab:meituan_result}, and Table \ref{tab:nc_results} \textit{all illustrate the performance of pre-training different DGNN encoders with \model is consistently better than the vanilla task-supervised pre-training strategy.}
This demonstrates both the effectiveness and generalization of the proposed \model method in different DGNN backbones.}



{\textbf{Inductive Generalization}. In order to further investigate the generalization of \model on inductive downstream tasks, we further compare the performance of \model with the JODIE encoder under the inductive link prediction task. 
The results in Table \ref{tab:ablation-three-transfer} demonstrate the performance of \model on three transfers under the inductive setting. We observe that \model significantly improves AUC under time transfer by about 10\% on the Entertainment and Outdoors datasets, respectively. Moreover, under the time+field transfer, the most challenging transfer, \model still achieves 5.17\% and 3.19\% AUC gains on Entertainment and Outdoors datasets, respectively.
Altogether, the results demonstrate that \model is also able to well generalize on the inductive downstream task.}

\begin{table}[tbp]
  \centering
  \vspace{-0.6em}
  \caption{Inductive study results on AUC and AP metrics. }
  \resizebox{0.955\linewidth}{!}{
    \begin{tabular}{cc|cc}
    \toprule
    \multicolumn{2}{c|}{Evaluation Metric} & AUC   & AP \\
    \midrule
    \multirow{4}[4]{*}{\rotatebox{90}{Beauty}} & No Pre-train & 0.6798±0.0106 & 0.6848±0.0220 \\
\cmidrule{2-4}          & \model (T) & \textbf{0.7219}±0.0023 \textbf{(+6.19\%)} & \textbf{0.7409}±0.0013 \textbf{(+8.19\%)} \\
          & \model (F) & 0.6983±0.0081 (+2.72\%) & 0.7088±0.0099 (+3.50\%) \\
          & \model (T+F) & 0.7026±0.0034 (+3.35\%) & 0.7201±0.0052 (+5.15\%) \\
    \midrule
    \multirow{4}[4]{*}{\rotatebox{90}{Luxury}} & No Pre-train & 0.6927±0.0038 & 0.6991±0.0050 \\
\cmidrule{2-4}          & \model (T) & \textbf{0.7187}±0.0023 \textbf{(+3.75\%)} & \textbf{0.7358}±0.0046 \textbf{(+5.25\%)} \\
          & \model (F) & 0.7100±0.0030 (+2.50\%) & 0.7267±0.0017 (+3.95\%) \\
          & \model (T+F) & 0.7059±0.0034 (+1.91\%) & 0.7241±0.0055 (+3.58\%) \\
    \midrule
    \multirow{4}[4]{*}{\rotatebox{90}{Entertain}} & No Pre-train & 0.7237±0.0130 & 0.7407±0.0115 \\
\cmidrule{2-4}          & \model (T) & \textbf{0.8015}±0.0049 \textbf{(+10.75\%)} & \textbf{0.8071}±0.0043 \textbf{(+8.96\%)} \\
          & \model (F) & 0.7737±0.0036 (+6.91\%) & 0.7792±0.0043 (+5.20\%) \\
          & \model (T+F) & 0.7611±0.0091 (+5.17\%) & 0.7714±0.0066 (+4.14\%) \\
    \midrule
    \multirow{4}[4]{*}{\rotatebox{90}{Outdoors}} & No Pre-train & 0.7079±0.0036 & 0.7294±0.0034 \\
\cmidrule{2-4}          & \model (T) & \textbf{0.7822}±0.0016 \textbf{(+10.50\%)} & \textbf{0.7980}±0.0010 \textbf{(+9.40\%)} \\
          & \model (F) & 0.7579±0.0025 (+7.06\%) & 0.7712±0.0029 (+5.73\%) \\
          & \model (T+F) & 0.7356±0.0150 (+3.91\%) & 0.7551±0.0117 (+3.52\%) \\
    \bottomrule
    \end{tabular}%
    }
    \vspace{-0.8em}
  \label{tab:ablation-three-transfer}%
\end{table}%



\subsection{Ablation Study}\label{exp:ablation}
    \begin{figure}[t]
    \centering
    \vspace{-0.15em}
    \includegraphics[width=0.915\linewidth]{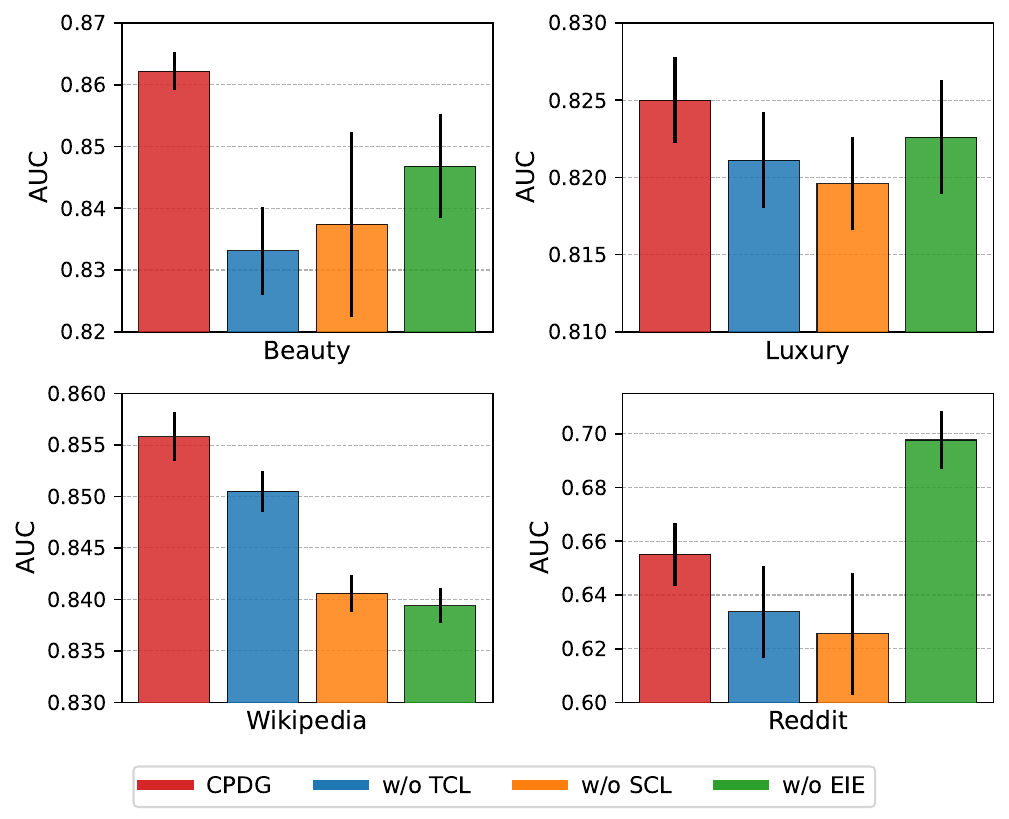}
    \caption{Ablation study results between \model and its three variants on Amazon-Beauty, Amazon-Luxury, {Wikipedia, and Reddit}.}
    \label{fig:ablation}
    \vspace{-1em}
\end{figure}

\begin{table}[t]
  \centering
  \vspace{-0.8em}
  \caption{Fine-tuning strategies comparison results on Amazon-Beauty and Amazon-Luxury, under \textit{time+field transfer} setting.}
    \resizebox{0.82\linewidth}{!}{
    \begin{tabular}{cc|cc}
    \toprule
    \multicolumn{2}{c|}{Evaluation Metric} & AUC   & AP \\
    \midrule
    \multirow{4}[4]{*}{\rotatebox{90}{Beauty}} & Full & 0.8468±0.0084 & 0.8423±0.0077 \\
\cmidrule{2-4}          & EIE-mean & 0.8496±0.0048 & 0.8440±0.0049 \\
          & EIE-attn & 0.8517±0.0031 & 0.8472±0.0026 \\
          & EIE-GRU & \textbf{0.8622}±0.0031 & \textbf{0.8541}±0.0033 \\
    \midrule
    \multirow{4}[4]{*}{\rotatebox{90}{Luxury}} & Full & 0.8226±0.0074 & 0.8213±0.0064 \\
\cmidrule{2-4}          & EIE-mean & 0.8237±0.0039 & 0.8244±0.0020 \\
          & EIE-attn & 0.8201±0.0113 & 0.8214±0.0080 \\
          & EIE-GRU & \textbf{0.8250}±0.0028 & \textbf{0.8250}±0.0034 \\
    \bottomrule
    \end{tabular}%
    }
  \label{tab:fine_tune}%
  \vspace{-1em}
\end{table}%

The proposed \model method consists of three key carefully designed modules, including temporal contrast (TC), structural contrast (SC), and evolution information enhanced (EIE) fine-tuning.
{To shed light on the effectiveness of these modules, we conduct ablation studies on the variants of \model on both dynamic link prediction datasets (Beauty and Luxury) and dynamic node classification datasets (Wikipedia and Reddit).}
As shown in Figure \ref{fig:ablation}, we report the results of \model without temporal contrast (w/o TC), without structural contrast (w/o SC), and without the evolution information enhanced fine-tuning (w/o EIE) of the Amazon Review dataset under the \textit{time+field transfer} setting. 

In general, we have the following findings:
{(i) \textit{The performance of \model is significantly superior to the \model variants without TC, SC, or EIE in general} (w/o EIE achieve the better result on Reddit), which indicates the effectiveness of all three modules of the proposed \model method.
(ii) Different datasets have different preferences for temporal and structural patterns.
The performance of \model w/o TC drops more than w/o SC on the Beauty field, indicating that temporal contrast provides more meaningful information for pre-training in this case. While the observation is just the opposite on the Luxury, Wikipedia, and Reddit fields. It suggests that different dynamic graphs have different emphases on temporal evolution information and structural evolution information, so it is meaningful for the \model to adopt both temporal and structural contrast collaboratively.}

\subsection{Discussion on EIE}\label{exp:eie}
In this subsection, we investigate the impact of the optional evolution information enhanced (EIE) fine-tuning strategy.
There are different types of $f_{EI}(\cdot)$ that can be derived from different EIE variants, such as EIE-mean, EIE-attn, and EIE-GRU, which implement $f_{EI}(\cdot)$ with mean pooling, attention mechanism~\cite{zhou2018deep}, and GRU operation, respectively. 

The results on two fields of the Amazon dataset under the \textit{time+field transfer} task with JODIE backbone are shown in Table \ref{tab:fine_tune}, from which we can observe that: 
{(i) \textbf{All the EIE variants generally perform better than the full fine-tuning strategy for downstream tasks}, even the simple mean pooling also has a good improvement compared to the full fine-tuning. This indicates that the information stored in pre-trained memory contains rich evolution information, which demonstrates the effectiveness of EIE for downstream tasks.  
(ii) \textbf{EIE-GRU performs best against other variants}. This observation demonstrates that the GRU operator has a better ability to capture the evolution pattern of pre-trained memory checkpoints. Due to the relatively high computational cost of GRU, the efficient EIE-mean can also be a suitable choice.}

\subsection{Parameter Study}\label{exp:param}
\begin{figure}[tbp]
    \centering
    \vspace{-0.8em}
    \includegraphics[width=0.90\linewidth, trim=0cm 0cm 0cm 0cm,clip]{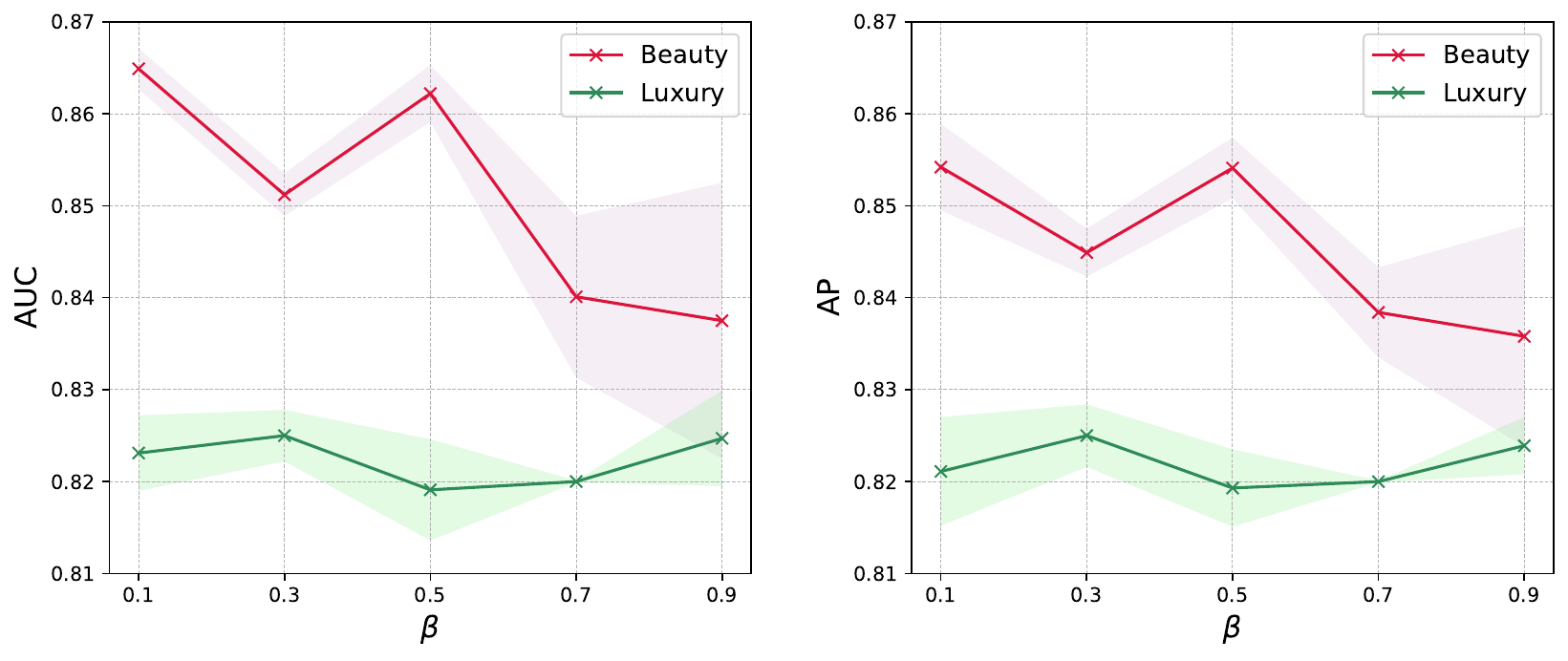}
    \caption{Hyper-parameter study of different $\beta$ value on Amazon-Beauty and Amazon-Luxury, under the \textit{time+field transfer} setting.}
    \vspace{-0.8em}
    \label{fig:beta}
\end{figure}

\textbf{Effect of the balanced parameter $\beta$}. We investigate the impact of hyper-parameter $\beta$ in Eq.(\ref{equ:beta}) and report the results in Figure \ref{fig:beta}. We can observe that the performance of Beauty drops with the $\beta$ increases in general, while that of Luxury is relatively stable. One possible reason is that temporal information is more essential in Beauty, while temporal information is almost as important as structural information in Luxury.

\begin{figure}[tbp]
    \centering
    \includegraphics[width=0.90\linewidth, trim=0cm 0cm 0cm 0cm,clip]{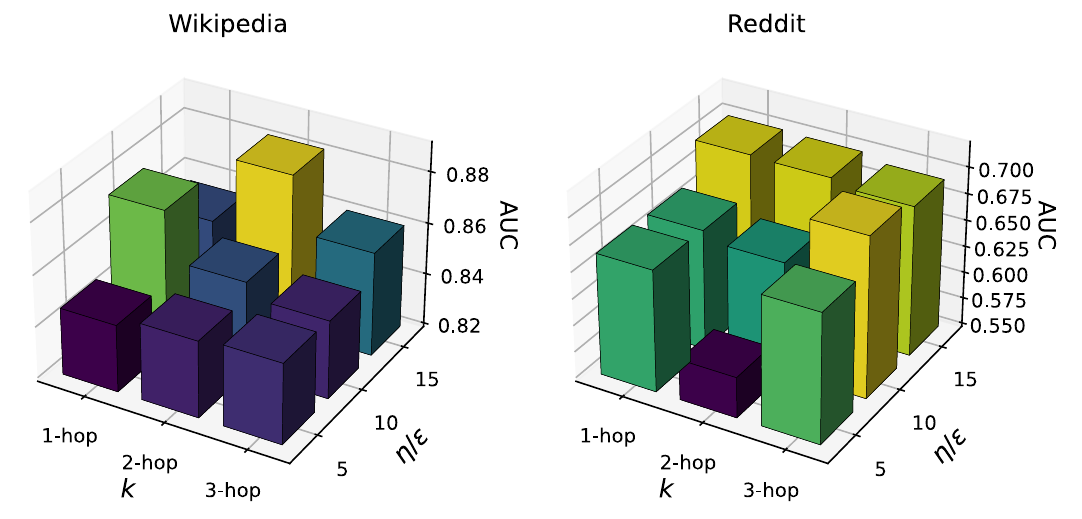}
    \caption{{Hyper-parameter study of different combination of $k$ and $\eta/\epsilon$.}}
    \label{fig:wide_depth}
    \vspace{-0.8em}
\end{figure}

{\textbf{Effect of contrastive subgraph parameters $\eta/\epsilon$ and $k$}. 
We investigate the combination effect of contrastive subgraph parameters $\eta/\epsilon$ and $k$ as illustrated in Figure \ref{fig:wide_depth}. We can observe that increasing $\eta/\epsilon$ (the width of the subgraph) generally has an improving impact on model performance, while deepening $k$ (the depth of the subgraph) may not necessarily bring a positive effect on the performance.}

\begin{figure}[!tbp]
    \centering
    \includegraphics[width=0.90\linewidth, trim=0cm 0cm 0cm 0cm,clip]{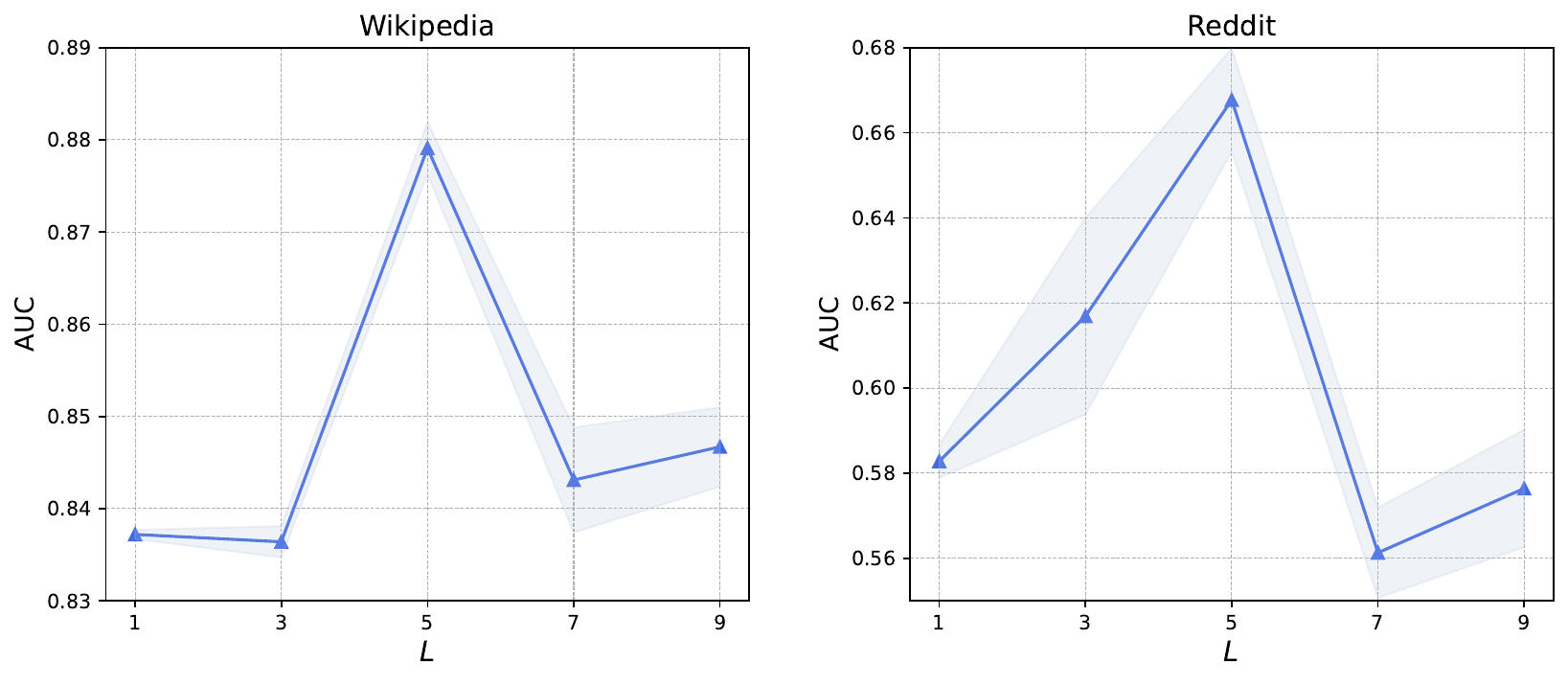}
    \caption{{Hyper-parameter study of different checkpoint length $L$ in EIE.}}
    \label{fig:seq_len}
    \vspace{-1.3em}
\end{figure}

{\textbf{Effect of checkpoint length $L$ in EIE}. We also explored the impact of $L$ by varying it from 1 to 9 with the step size of 2 as shown in Figure \ref{fig:seq_len}. The results reveal that a checkpoint length of 5 yields relatively good performance while utilizing a length that is either too small or too large does not provide optimal benefits.}


%% file: conclusion.tex
In this paper, we propose to address the difficulties of applying DGNNs in industrial scenarios.
We propose a novel \model method to capture the long-short term temporal evolution patterns and discriminative structural patterns through flexible structural-temporal subgraph samplers along with structural-temporal contrastive pre-training schemes.
Furthermore, we introduce an optional evolution information enhanced fine-tuning strategy to take advantage of the evolved patterns during pre-training. Extensive experiments on widely used dynamic graph datasets and an industrial dataset in Meituan demonstrate the effectiveness of our proposed \model method. 
In future works, we will explore the \model in more practical scenarios, such as industrial recommender systems.

%% file: main.bbl
\begin{thebibliography}{10}
\providecommand{\url}[1]{#1}
\csname url@samestyle\endcsname
\providecommand{\newblock}{\relax}
\providecommand{\bibinfo}[2]{#2}
\providecommand{\BIBentrySTDinterwordspacing}{\spaceskip=0pt\relax}
\providecommand{\BIBentryALTinterwordstretchfactor}{4}
\providecommand{\BIBentryALTinterwordspacing}{\spaceskip=\fontdimen2\font plus
\BIBentryALTinterwordstretchfactor\fontdimen3\font minus
  \fontdimen4\font\relax}
\providecommand{\BIBforeignlanguage}[2]{{%
\expandafter\ifx\csname l@#1\endcsname\relax
\typeout{** WARNING: IEEEtran.bst: No hyphenation pattern has been}%
\typeout{** loaded for the language `#1'. Using the pattern for}%
\typeout{** the default language instead.}%
\else
\language=\csname l@#1\endcsname
\fi
#2}}
\providecommand{\BIBdecl}{\relax}
\BIBdecl

\bibitem{ying2018graph}
R.~Ying, R.~He, K.~Chen, P.~Eksombatchai, W.~L. Hamilton, and J.~Leskovec,
  ``Graph convolutional neural networks for web-scale recommender systems,'' in
  \emph{Proceedings of the 24th ACM SIGKDD Conference on Knowledge Discovery
  and Data Mining}, 2018, pp. 974--983.

\bibitem{wu2020comprehensive}
Z.~Wu, S.~Pan, F.~Chen, G.~Long, C.~Zhang, and S.~Y. Philip, ``A comprehensive
  survey on graph neural networks,'' \emph{IEEE transactions on neural networks
  and learning systems}, vol.~32, no.~1, pp. 4--24, 2020.

\bibitem{bei2023reinforcement}
Y.~Bei, S.~Zhou, Q.~Tan, H.~Xu, H.~Chen, Z.~Li, and J.~Bu, ``Reinforcement
  neighborhood selection for unsupervised graph anomaly detection,'' in
  \emph{2023 IEEE International Conference on Data Mining (ICDM)}, 2023.

\bibitem{pastor2001epidemic}
R.~Pastor-Satorras and A.~Vespignani, ``Epidemic spreading in scale-free
  networks,'' \emph{Physical review letters}, vol.~86, no.~14, p. 3200, 2001.

\bibitem{may2001infection}
R.~M. May and A.~L. Lloyd, ``Infection dynamics on scale-free networks,''
  \emph{Physical Review E}, vol.~64, no.~6, p. 066112, 2001.

\bibitem{li2019deep}
Y.~Li, C.~Huang, L.~Ding, Z.~Li, Y.~Pan, and X.~Gao, ``Deep learning in
  bioinformatics: Introduction, application, and perspective in the big data
  era,'' \emph{Methods}, vol. 166, pp. 4--21, 2019.

\bibitem{jumper2021highly}
J.~Jumper, R.~Evans, A.~Pritzel, T.~Green, M.~Figurnov, O.~Ronneberger,
  K.~Tunyasuvunakool, R.~Bates, A.~{\v{Z}}{\i}dek, A.~Potapenko \emph{et~al.},
  ``Highly accurate protein structure prediction with alphafold,''
  \emph{Nature}, vol. 596, no. 7873, pp. 583--589, 2021.

\bibitem{wu2020graph}
S.~Wu, F.~Sun, W.~Zhang, X.~Xie, and B.~Cui, ``Graph neural networks in
  recommender systems: a survey,'' \emph{ACM Computing Surveys}, vol.~55,
  no.~5, pp. 1--37, 2022.

\bibitem{he2020lightgcn}
X.~He, K.~Deng, X.~Wang, Y.~Li, Y.~Zhang, and M.~Wang, ``Lightgcn: Simplifying
  and powering graph convolution network for recommendation,'' in
  \emph{Proceedings of the 43rd International ACM SIGIR conference on research
  and development in Information Retrieval}, 2020, pp. 639--648.

\bibitem{bei2023non}
Y.~Bei, H.~Chen, S.~Chen, X.~Huang, S.~Zhou, and F.~Huang, ``Non-recursive
  cluster-scale graph interacted model for click-through rate prediction,'' in
  \emph{Proceedings of the 32nd ACM International Conference on Information and
  Knowledge Management}, 2023, pp. 3748--3752.

\bibitem{skarding2021foundations}
J.~Skarding, B.~Gabrys, and K.~Musial, ``Foundations and modeling of dynamic
  networks using dynamic graph neural networks: A survey,'' \emph{IEEE Access},
  vol.~9, pp. 79\,143--79\,168, 2021.

\bibitem{xu2020inductive}
D.~Xu, C.~Ruan, E.~K{\"{o}}rpeoglu, S.~Kumar, and K.~Achan, ``Proceedings of
  the inductive representation learning on temporal graphs,'' in
  \emph{International Conference on Learning Representations}, 2020.

\bibitem{nguyen2018continuous}
G.~H. Nguyen, J.~B. Lee, R.~A. Rossi, N.~K. Ahmed, E.~Koh, and S.~Kim,
  ``Continuous-time dynamic network embeddings,'' in \emph{Companion
  proceedings of the web conference 2018}, 2018, pp. 969--976.

\bibitem{kumar2018learning}
S.~Kumar, X.~Zhang, and J.~Leskovec, ``Predicting dynamic embedding trajectory
  in temporal interaction networks,'' in \emph{Proceedings of the 25th ACM
  SIGKDD Conference on Knowledge Discovery and Data Mining}, 2019, pp.
  1269--1278.

\bibitem{trivedi2019dyrep}
R.~Trivedi, M.~Farajtabar, P.~Biswal, and H.~Zha, ``Dyrep: Learning
  representations over dynamic graphs,'' in \emph{International Conference on
  Learning Representations}, 2019.

\bibitem{rossi2020temporal}
E.~Rossi, B.~Chamberlain, F.~Frasca, D.~Eynard, F.~Monti, and M.~Bronstein,
  ``Temporal graph networks for deep learning on dynamic graphs,'' in
  \emph{ICML 2020 Workshop on Graph Representation Learning}, 2020.

\bibitem{you2022roland}
J.~You, T.~Du, and J.~Leskovec, ``Roland: graph learning framework for dynamic
  graphs,'' in \emph{Proceedings of the 28th ACM SIGKDD Conference on Knowledge
  Discovery and Data Mining}, 2022, pp. 2358--2366.

\bibitem{devlin2019bert}
J.~Devlin, M.-W. Chang, K.~Lee, and K.~Toutanova, ``Bert: Pre-training of deep
  bidirectional transformers for language understanding,'' 2019, pp.
  4171--4186.

\bibitem{khan2022transformers}
S.~Khan, M.~Naseer, M.~Hayat, S.~W. Zamir, F.~S. Khan, and M.~Shah,
  ``Transformers in vision: A survey,'' \emph{ACM computing surveys (CSUR)},
  vol.~54, no. 10s, pp. 1--41, 2022.

\bibitem{hu2019strategies}
W.~Hu, B.~Liu, J.~Gomes, M.~Zitnik, P.~Liang, V.~S. Pande, and J.~Leskovec,
  ``Strategies for pre-training graph neural networks,'' in \emph{International
  Conference on Learning Representations}, 2020.

\bibitem{qiu2020gcc}
J.~Qiu, Q.~Chen, Y.~Dong, J.~Zhang, H.~Yang, M.~Ding, K.~Wang, and J.~Tang,
  ``Gcc: Graph contrastive coding for graph neural network pre-training,'' in
  \emph{Proceedings of the 26th ACM SIGKDD Conference on Knowledge Discovery
  and Data Mining}, 2020, pp. 1150--1160.

\bibitem{hu2020gpt}
Z.~Hu, Y.~Dong, K.~Wang, K.-W. Chang, and Y.~Sun, ``Gpt-gnn: Generative
  pre-training of graph neural networks,'' in \emph{Proceedings of the 26th ACM
  SIGKDD Conference on Knowledge Discovery and Data Mining}, 2020, pp.
  1857--1867.

\bibitem{lu2021learning}
Y.~Lu, X.~Jiang, Y.~Fang, and C.~Shi, ``Learning to pre-train graph neural
  networks,'' in \emph{Proceedings of the AAAI conference on artificial
  intelligence}, vol.~35, no.~5, 2021, pp. 4276--4284.

\bibitem{sun2022gppt}
M.~Sun, K.~Zhou, X.~He, Y.~Wang, and X.~Wang, ``Gppt: Graph pre-training and
  prompt tuning to generalize graph neural networks,'' in \emph{Proceedings of
  the 28th ACM SIGKDD Conference on Knowledge Discovery and Data Mining}, 2022,
  pp. 1717--1727.

\bibitem{liu2023graphprompt}
Z.~Liu, X.~Yu, Y.~Fang, and X.~Zhang, ``Graphprompt: Unifying pre-training and
  downstream tasks for graph neural networks,'' in \emph{Proceedings of the ACM
  Web Conference 2023}, 2023, pp. 417--428.

\bibitem{tian2021self}
S.~Tian, R.~Wu, L.~Shi, L.~Zhu, and T.~Xiong, ``Self-supervised representation
  learning on dynamic graphs,'' in \emph{Proceedings of the 30th ACM
  International Conference on Information \& Knowledge Management}, 2021, pp.
  1814--1823.

\bibitem{velickovic2019deep}
P.~Velickovic, W.~Fedus, W.~L. Hamilton, P.~Li{\`{o}}, Y.~Bengio, and R.~D.
  Hjelm, ``Deep graph infomax,'' in \emph{International Conference on Learning
  Representations}, 2019.

\bibitem{chen2022pre}
K.-J. Chen, J.~Zhang, L.~Jiang, Y.~Wang, and Y.~Dai, ``Pre-training on dynamic
  graph neural networks,'' \emph{Neurocomputing}, vol. 500, pp. 679--687, 2022.

\bibitem{sun2022self}
L.~Sun, J.~Ye, H.~Peng, and P.~S. Yu, ``A self-supervised riemannian gnn with
  time varying curvature for temporal graph learning,'' in \emph{Proceedings of
  the 31st ACM International Conference on Information \& Knowledge
  Management}, 2022, pp. 1827--1836.

\bibitem{zhang2020deep}
Z.~Zhang, P.~Cui, and W.~Zhu, ``Deep learning on graphs: A survey,'' \emph{IEEE
  Transactions on Knowledge and Data Engineering}, 2020.

\bibitem{kipf2016semi}
T.~N. Kipf and M.~Welling, ``Semi-supervised classification with graph
  convolutional networks,'' 2017.

\bibitem{hamilton2017inductive}
W.~Hamilton, Z.~Ying, and J.~Leskovec, ``Inductive representation learning on
  large graphs,'' in \emph{Advances in Neural Information Processing Systems},
  vol.~30, 2017.

\bibitem{velivckovic2018graph}
P.~Velickovic, G.~Cucurull, A.~R. Arantxa~Casanova, P.~Lio, and Y.~Bengio,
  ``Graph attention networks,'' in \emph{International Conference on Learning
  Representations}, 2018.

\bibitem{zhang2023alleviating}
Y.~Zhang, Y.~Bei, S.~Yang, H.~Chen, Z.~Li, L.~Chen, and F.~Huang, ``Alleviating
  behavior data imbalance for multi-behavior graph collaborative filtering,''
  in \emph{2023 IEEE International Conference on Data Mining Workshops
  (ICDMW)}, 2023.

\bibitem{wang2020inductive}
Y.~Wang, Y.-Y. Chang, Y.~Liu, J.~Leskovec, and P.~Li, ``Inductive
  representation learning in temporal networks via causal anonymous walks,'' in
  \emph{International Conference on Learning Representations}, 2020.

\bibitem{jiang2021pre}
X.~Jiang, T.~Jia, Y.~Fang, C.~Shi, Z.~Lin, and H.~Wang, ``Pre-training on
  large-scale heterogeneous graph,'' in \emph{Proceedings of the 27th ACM
  SIGKDD Conference on Knowledge Discovery and Data Mining}, 2021, pp.
  756--766.

\bibitem{lichtenwalter2010new}
R.~N. Lichtenwalter, J.~T. Lussier, and N.~V. Chawla, ``New perspectives and
  methods in link prediction,'' in \emph{Proceedings of the 16th ACM SIGKDD
  Conference on Knowledge Discovery and Data Mining}, 2010, pp. 243--252.

\bibitem{medsker2001recurrent}
L.~R. Medsker and L.~Jain, ``Recurrent neural networks,'' \emph{Design and
  Applications}, vol.~5, pp. 64--67, 2001.

\bibitem{hochreiter1997long}
S.~Hochreiter and J.~Schmidhuber, ``Long short-term memory,'' \emph{Neural
  computation}, vol.~9, no.~8, pp. 1735--1780, 1997.

\bibitem{chung2014empirical}
J.~Chung, C.~Gulcehre, K.~Cho, and Y.~Bengio, ``Empirical evaluation of gated
  recurrent neural networks on sequence modeling,'' in \emph{NIPS 2014 Workshop
  on Deep Learning}, 2014.

\bibitem{yu2019adaptive}
Z.~Yu, J.~Lian, A.~Mahmoody, G.~Liu, and X.~Xie, ``Adaptive user modeling with
  long and short-term preferences for personalized recommendation.'' in
  \emph{Proceedings of the Twenty-Eighth International Joint Conference on
  Artificial Intelligence, {IJCAI-19}}, 2019, pp. 4213--4219.

\bibitem{chi2023modeling}
H.~Chi, H.~Xu, M.~Liu, Y.~Bei, S.~Zhou, D.~Liu, and M.~Zhang, ``Modeling
  spatiotemporal periodicity and collaborative signal for local-life service
  recommendation,'' \emph{arXiv preprint arXiv:2309.12565}, 2023.

\bibitem{sun2022graph}
H.~Sun, G.~Yu, P.~Zhang, B.~Zhang, X.~Wang, and D.~Wang, ``Graph based
  long-term and short-term interest model for click-through rate prediction,''
  in \emph{Proceedings of the 31st ACM International Conference on Information
  \& Knowledge Management}, 2022, pp. 1818--1826.

\bibitem{perozzi2014deepwalk}
B.~Perozzi, R.~Al-Rfou, and S.~Skiena, ``Deepwalk: Online learning of social
  representations,'' in \emph{Proceedings of the 20th ACM SIGKDD Conference on
  Knowledge Discovery and Data Mining}, 2014, pp. 701--710.

\bibitem{ying2018hierarchical}
Z.~Ying, J.~You, C.~Morris, X.~Ren, W.~Hamilton, and J.~Leskovec,
  ``Hierarchical graph representation learning with differentiable pooling,''
  in \emph{Advances in Neural Information Processing Systems}, vol.~31, 2018.

\bibitem{balntas2016learning}
D.~P. Vassileios~Balntas, Edgar~Riba and K.~Mikolajczyk, ``Learning local
  feature descriptors with triplets and shallow convolutional neural
  networks,'' in \emph{Proceedings of the British Machine Vision Conference},
  2016.

\bibitem{ruby2020binary}
U.~Ruby and V.~Yendapalli, ``Binary cross entropy with deep learning technique
  for image classification,'' \emph{Int. J. Adv. Trends Comput. Sci. Eng},
  vol.~9, no.~10, 2020.

\bibitem{burch1993monte}
R.~Burch, F.~N. Najm, P.~Yang, and T.~N. Trick, ``A monte carlo approach for
  power estimation,'' \emph{IEEE Transactions on Very Large Scale Integration
  (VLSI) Systems}, vol.~1, no.~1, pp. 63--71, 1993.

\bibitem{ni2019justifying}
J.~Ni, J.~Li, and J.~McAuley, ``Justifying recommendations using
  distantly-labeled reviews and fine-grained aspects,'' in \emph{Proceedings of
  EMNLP-IJCNLP}, 2019, pp. 188--197.

\bibitem{liu2014exploiting}
Y.~Liu, W.~Wei, A.~Sun, and C.~Miao, ``Exploiting geographical neighborhood
  characteristics for location recommendation,'' in \emph{Proceedings of the
  23rd ACM International Conference on Information \& Knowledge Management},
  2014, pp. 739--748.

\bibitem{xu2018powerful}
K.~Xu, W.~Hu, J.~Leskovec, and S.~Jegelka, ``How powerful are graph neural
  networks?'' in \emph{International Conference on Learning Representations},
  2019.

\bibitem{hou2022graphmae}
Z.~Hou, X.~Liu, Y.~Cen, Y.~Dong, H.~Yang, C.~Wang, and J.~Tang, ``Graphmae:
  Self-supervised masked graph autoencoders,'' in \emph{Proceedings of the 28th
  ACM SIGKDD Conference on Knowledge Discovery and Data Mining}, 2022, pp.
  594--604.

\bibitem{zhou2018deep}
G.~Zhou, X.~Zhu, C.~Song, Y.~Fan, H.~Zhu, X.~Ma, Y.~Yan, J.~Jin, H.~Li, and
  K.~Gai, ``Deep interest network for click-through rate prediction,'' in
  \emph{Proceedings of the 24th ACM SIGKDD Conference on Knowledge Discovery
  and Data Mining}, 2018, pp. 1059--1068.

\end{thebibliography}
